\documentclass{article}

\usepackage{arxiv}

\usepackage[utf8]{inputenc} 
\usepackage[T1]{fontenc}    
\usepackage{hyperref}       
\usepackage{url}            
\usepackage{booktabs}       
\usepackage{amsfonts}       
\usepackage{nicefrac}       
\usepackage{microtype}      
\usepackage{cleveref}       
\usepackage{lipsum}         
\usepackage{graphicx}
\usepackage{natbib}
\usepackage{doi}

\title{Temporal vs. Spatial: Comparing DINOv3 and V-JEPA2 Feature Representations for Video Action Analysis}

\date{}

\author{ \href{https://orcid.org/0009-0002-3124-633X}{\includegraphics[scale=0.06]{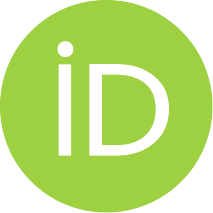}\hspace{1mm}Sai Varun Kodathala} \\
	Research and Development\\
	Sports Vision, Inc.\\
	Minnetonka, MN \\
	\texttt{varun@sportsvision.ai} \\
	\And
	Rakesh Vunnam\\
	Research and Development\\
	Vizworld, Inc.\\
	Minnetonka, MN \\
	\texttt{rakesh@vizworld.ai} \\
}

\renewcommand{\shorttitle}{\textit{Sports Vision's} Research}

\hypersetup{
pdftitle={A template for the arxiv style},
pdfsubject={q-bio.NC, q-bio.QM},
pdfauthor={David S.~Hippocampus, Elias D.~Striatum},
pdfkeywords={First keyword, Second keyword, More},
}

\begin{document}
\maketitle

\begin{abstract}
This study presents a comprehensive comparative analysis of two prominent self-supervised learning architectures for video action recognition: DINOv3, which processes frames independently through spatial feature extraction, and V-JEPA2, which employs joint temporal modeling across video sequences. We evaluate both approaches on the UCF Sports dataset, examining feature quality through multiple dimensions including classification accuracy, clustering performance, intra-class consistency, and inter-class discrimination. Our analysis reveals fundamental architectural trade-offs: DINOv3 achieves superior clustering performance (Silhouette score: 0.31 vs 0.21) and demonstrates exceptional discrimination capability (6.16× separation ratio) particularly for pose-identifiable actions, while V-JEPA2 exhibits consistent reliability across all action types with significantly lower performance variance ($\sigma$ = 0.094 vs 0.288). Through action-specific evaluation, we identify that DINOv3's spatial processing architecture excels at static pose recognition but shows degraded performance on motion-dependent actions, whereas V-JEPA2's temporal modeling provides balanced representation quality across diverse action categories. These findings contribute to the understanding of architectural design choices in video analysis systems and provide empirical guidance for selecting appropriate feature extraction methods based on task requirements and reliability constraints.

\end{abstract}

\section{Introduction}

Video understanding represents one of the most fundamental challenges in computer vision, demanding sophisticated approaches that can capture both spatial visual content and temporal dynamics inherent in sequential data~\cite{aggarwal2011human,moeslund2006survey}. The emergence of self-supervised learning has revolutionized this domain by enabling models to learn rich representations from unlabeled video data, circumventing the expensive annotation requirements that have traditionally limited scalable video analysis~\cite{schiappa2023self}. However, a fundamental architectural question persists in the field: whether independent frame-level processing or joint temporal modeling produces superior action representations for downstream video understanding tasks.

Contemporary self-supervised video representation learning approaches generally fall into two distinct paradigms. Frame-based methods process video frames independently, leveraging powerful image encoders to extract detailed spatial features from each frame before aggregating these representations for sequence-level understanding~\cite{wang2016temporal,krizhevsky2012imagenet}. Sequence-based methods jointly process temporal sequences, explicitly modeling frame relationships and motion dynamics through specialized architectures designed to capture spatiotemporal dependencies~\cite{tran2015learning,carreira2017quo}.

Recent advances in self-supervised learning have produced increasingly sophisticated models within both paradigms. DINOv3, representing the latest evolution in self-supervised visual representation learning, employs a teacher-student distillation framework trained on large-scale image collections to learn dense visual features without supervision~\cite{metaai2024dinov3}. When applied to video understanding, DINOv3 processes each frame independently, producing high-dimensional embeddings that capture rich spatial details, object relationships, and visual semantics within individual frames. This approach maximizes the utilization of powerful image representations while maintaining computational efficiency through parallel frame processing.

Conversely, V-JEPA2 (Video Joint Embedding Predictive Architecture 2) represents a fundamentally different approach that jointly processes video frames to generate temporal tokens encoding motion dynamics and frame relationships~\cite{assran2025vjepa2}. Built upon the Joint Embedding Predictive Architecture framework~\cite{lecun2022path}, V-JEPA2 employs masked prediction objectives to learn spatiotemporal representations that explicitly model temporal dependencies, motion patterns, and cross-frame relationships. This architecture prioritizes temporal coherence and dynamic understanding over frame-level spatial detail.

The architectural differences between these approaches reflect a deeper tension in video representation learning between spatial fidelity and temporal modeling~\cite{dosovitskiy2021image,caron2021emerging}. Frame-based methods like DINOv3 excel at capturing fine-grained visual details, object attributes, and complex spatial relationships within individual frames, potentially enabling superior performance on tasks requiring detailed visual understanding. However, these methods may struggle to capture subtle motion patterns, temporal dependencies, and dynamic interactions that unfold across multiple frames. Sequence-based methods like V-JEPA2 prioritize temporal coherence and motion understanding, potentially excelling at tasks requiring temporal reasoning and dynamic scene understanding, but may sacrifice spatial detail in favor of temporal compression.

This fundamental trade-off becomes particularly pronounced in action recognition scenarios where both spatial configuration and temporal dynamics contribute to accurate classification. Athletic movements, for instance, require understanding of both precise body positioning (spatial) and movement trajectories (temporal). Similarly, complex actions involving object manipulation demand both detailed object recognition and temporal sequence understanding. The optimal balance between these complementary sources of information remains an open research question.

To address this knowledge gap, we conduct a comprehensive empirical analysis comparing DINOv3 and V-JEPA2 feature representations across multiple dimensions of video understanding performance. Our investigation employs a systematic evaluation framework that assesses both approaches across intra-class consistency, inter-class separation, viewpoint invariance, temporal smoothness, and motion sensitivity. We utilize the UCF Sports dataset~\cite{soomro2014action} as our primary evaluation benchmark, employing keyframe selection methodology to ensure fair comparison by providing both models with identical temporal sampling.

Our contributions include: (1) the first systematic comparison of frame-based versus sequence-based self-supervised video representations on sports action recognition tasks, (2) a comprehensive evaluation framework that assesses multiple dimensions of video understanding performance beyond traditional accuracy metrics, (3) empirical evidence characterizing when each approach excels, providing practical guidance for model selection based on application requirements, and (4) insights into the fundamental trade-offs between spatial detail preservation and temporal context modeling in contemporary self-supervised video understanding systems.

Rather than advocating for the universal superiority of either approach, this work seeks to characterize the complementary strengths and limitations of frame-based and sequence-based video representation learning. Through systematic analysis, we provide practitioners with empirically-grounded guidance for selecting appropriate architectures based on their specific temporal complexity requirements and computational constraints.

\section{Related Work}

\subsection{Self-Supervised Video Representation Learning}

Self-supervised learning has emerged as a dominant paradigm for video representation learning, addressing the scalability limitations imposed by manual annotation requirements~\cite{donahue2015long,pathak2017curiosity}. Early approaches focused on pretext tasks specifically designed for temporal data, including frame order prediction~\cite{lee2017unsupervised}, clip rotation recognition, and temporal jigsaw puzzles~\cite{kim2019self}. These methods demonstrated that temporal structure provides rich supervisory signals for learning meaningful video representations without human-provided labels.

The evolution of self-supervised video learning can be broadly categorized into four methodological approaches: pretext task-based methods, generative approaches, contrastive learning, and cross-modal alignment~\cite{schiappa2022self}. Pretext task-based methods exploit temporal relationships through tasks such as frame sequence prediction~\cite{misra2016shuffle}, temporal order verification~\cite{buchler2018improving}, and motion-based puzzles. Generative approaches learn representations by reconstructing future frames or predicting masked portions of video sequences~\cite{vondrick2016generating,srivastava2015unsupervised}. Contrastive learning methods maximize agreement between different views of the same video while minimizing similarity to other videos~\cite{han2020self}. Cross-modal alignment approaches leverage relationships between video and other modalities such as audio or text~\cite{tian2019contrastive,arandjelovic2017look,owens2018audio}.

Recent advances have focused on scaling these approaches to larger datasets and more sophisticated architectures. The Video Joint Embedding Predictive Architecture (V-JEPA) framework introduced by Bardes et al.~\cite{bardes2024vjepa} represents a significant advancement in predictive self-supervised learning for video. This approach learns representations by predicting masked regions in abstract representation space rather than raw pixels, enabling more efficient training and better generalization. V-JEPA2 extends this framework with improved architectures and training procedures~\cite{assran2025vjepa2}.

\subsection{Frame-Based vs. Sequence-Based Processing}

The fundamental distinction between frame-based and sequence-based video processing has been a persistent theme in computer vision research~\cite{wang2016temporal,simonyan2014two}. Frame-based approaches treat videos as collections of independent images, leveraging powerful image understanding models to extract spatial features before temporal aggregation. This paradigm benefits from the rapid advances in image representation learning, particularly self-supervised methods like DINO and its successors~\cite{caron2021emerging,chen2021exploring}.

DINOv2 introduced significant improvements in self-supervised image representation learning through advanced distillation techniques and large-scale training~\cite{oquab2023dinov2}. The subsequent DINOv3 scales this approach further, training on large-scale image collections with sophisticated data curation and model architectures~\cite{metaai2024dinov3}. When applied to video understanding, these models process each frame independently, relying on subsequent temporal aggregation mechanisms to capture sequential information.

Sequence-based approaches explicitly model temporal dependencies through architectures designed for sequential data processing~\cite{bertasius2021space,arnab2021vivit}. Early work in this domain employed recurrent neural networks and temporal convolutional networks to capture motion patterns and temporal relationships~\cite{donahue2015long,feichtenhofer2016spatiotemporal}. More recent approaches leverage attention mechanisms and transformer architectures specifically adapted for spatiotemporal modeling~\cite{bertasius2021space,arnab2021vivit}.

The trade-off between spatial detail and temporal modeling has been empirically studied in various contexts. Bertasius et al.~\cite{bertasius2021space} demonstrated that pure attention-based approaches can achieve competitive performance with 3D convolutional networks while offering greater computational efficiency. However, these studies typically focus on supervised learning scenarios and do not comprehensively address the fundamental representational differences between frame-based and sequence-based self-supervised approaches.

\subsection{Temporal vs. Spatial Information in Video Understanding}

The relationship between spatial and temporal information in video understanding has been extensively studied, revealing complex interdependencies that challenge simplistic architectural choices~\cite{benyosef2021minimal}. Research has demonstrated that spatial information can compensate for temporal deficiencies and vice versa, suggesting that optimal video understanding systems must balance both dimensions~\cite{huang2018analyzing}.

Recent work in self-supervised video learning has begun to address these temporal-spatial trade-offs more directly. Co-training approaches exploit the complementary information between RGB frames and optical flow to improve temporal understanding~\cite{han2020self}. Other methods introduce ranking-based transformation recognition to learn temporal representations through relative temporal ordering~\cite{duan2022transrank}. These approaches suggest that effective video understanding requires sophisticated integration of spatial and temporal information streams.

\subsection{Action Recognition and Sports Video Analysis}

Action recognition represents one of the most challenging applications of video understanding, requiring sophisticated integration of spatial and temporal information~\cite{thomas2017computer,aggarwal2011human}. Sports video analysis presents particular challenges due to rapid motion, complex backgrounds, and the need to discriminate between subtly different actions~\cite{cioppa2020context,shao2020finegym}.

The UCF Sports dataset~\cite{rodriguez2008action} has become a standard benchmark for action recognition research. Comprising video sequences across multiple sport categories, the dataset provides realistic action recognition challenges with natural variations in viewpoint, background, and execution style. Unlike staged action datasets~\cite{soomro2012ucf101,kuehne2011hmdb}, UCF Sports captures authentic sports footage from broadcast sources, providing more representative evaluation conditions for real-world deployment scenarios.

Recent advances in sports video analysis have emphasized the importance of temporal modeling for capturing complex athletic movements~\cite{yu2018fine,qi2019sports,xu2022finediving}. The multi-phase nature of athletic actions requires representations that can capture both fine-grained spatial details and temporal dynamics. This domain provides an ideal testbed for comparing frame-based and sequence-based representation learning approaches.

\subsection{Keyframe Selection and Temporal Sampling}

Effective temporal sampling represents a critical component of video understanding systems, particularly for computationally constrained scenarios~\cite{zhu2018hidden,zhang2016real}. Traditional uniform sampling approaches may miss critical temporal events or include redundant information, motivating more sophisticated sampling strategies~\cite{vala2021key,tint2020key}.

Discrete Wavelet Transform (DWT) based approaches have shown promise for adaptive keyframe selection by identifying frames with significant temporal variation. The method introduced by Kodathala et al.~\cite{kodathala2025sv33bsportsvideounderstanding} employs DWT analysis combined with VGG16 feature extraction and Linear Discriminant Analysis to select 16 representative frames that capture critical temporal transitions. This approach ensures that both spatial content and temporal dynamics are preserved in the selected keyframes, enabling fair comparison between different representation learning approaches.

Alternative keyframe selection strategies include content-based methods that identify frames with maximum visual information~\cite{kapre2023improved}, motion-based approaches that prioritize frames with significant optical flow~\cite{singh2025key}, and clustering-based methods that ensure diverse temporal coverage~\cite{zhuang1998adaptive}. However, the DWT-based approach provides a principled framework that balances computational efficiency with temporal representativeness, making it particularly suitable for comparative analysis of video representation methods.

\subsection{Evaluation Methodologies for Video Representations}

Traditional evaluation of video representations has focused primarily on downstream task performance, particularly action recognition accuracy~\cite{karpathy2014large,carreira2017quo}. However, this approach provides limited insight into the fundamental characteristics of learned representations and may not reveal important differences in spatial versus temporal modeling capabilities~\cite{ng2015beyond}.

More comprehensive evaluation frameworks have been proposed to assess multiple dimensions of video understanding performance~\cite{hara2018can,wang2016temporal,feichtenhofer2019slowfast}. These include analysis of temporal consistency, spatial detail preservation, viewpoint invariance, and motion sensitivity. Such multi-dimensional evaluation approaches provide deeper insights into the strengths and limitations of different representation learning methods.

Recent work has emphasized the importance of analyzing representation quality beyond downstream performance metrics~\cite{maaten2008visualizing,alemi2017deep}. Visualization techniques, nearest neighbor analysis, and linear probing experiments can reveal important differences in representation structure and semantic organization~\cite{tran2018closer,wang2018non}. These analytical approaches are particularly valuable for understanding the fundamental trade-offs between different architectural choices in video representation learning.

\subsection{Gap in Current Literature}

Despite extensive research in both frame-based and sequence-based video representation learning, systematic comparative analysis of these approaches remains limited. Most existing work focuses on improving performance within a single paradigm rather than understanding the fundamental trade-offs between different architectural choices~\cite{laptev2005space,tian2019contrastive,chen2020simple}. Furthermore, existing comparisons typically rely on downstream task performance alone, providing insufficient insight into the underlying representational differences.

The lack of comprehensive comparative analysis is particularly pronounced in the self-supervised learning domain, where recent advances have produced sophisticated models within both paradigms but limited understanding of their relative strengths and limitations. This work addresses this gap by providing systematic empirical analysis of frame-based versus sequence-based self-supervised video representation learning, offering insights essential for informed architectural choices in video understanding applications.

\section{Methodology}

This section presents our comprehensive experimental framework for comparing DINOv3 and V-JEPA2 feature representations in sports video action analysis. Our methodology encompasses dataset preparation, keyframe extraction, feature extraction protocols, and multi-dimensional evaluation frameworks designed to assess both spatial and temporal modeling capabilities across diverse athletic movements.

\subsection{Dataset and Experimental Setup}

\subsubsection{UCF Sports Dataset}
We utilize the UCF Sports dataset \citep{soomro2014action} as our primary evaluation benchmark, which contains realistic sports action sequences captured from broadcast television. The dataset comprises video sequences across 13 distinct action classes with viewpoint variations: Diving-Side, Golf-Swing-Back, Golf-Swing-Front, Golf-Swing-Side, Kicking-Front, Kicking-Side, Lifting, Riding-Horse, Run-Side, SkateBoarding-Front, Swing-Bench, Swing-SideAngle, and Walk-Front. Each video contains natural variations in viewpoint, background complexity, lighting conditions, and execution style, making it particularly suitable for evaluating representation learning approaches under realistic deployment conditions.

The dataset presents several unique challenges that distinguish it from staged action recognition datasets. First, the broadcast nature of the videos introduces camera motion, zoom variations, and multiple viewpoints within single sequences, as evidenced by the viewpoint-specific class labels (e.g., Golf-Swing-Back vs. Golf-Swing-Front). Second, the athletic movements exhibit rapid temporal dynamics with critical biomechanical transitions occurring within brief time windows. Third, the diversity of sports actions requires models to capture both pose-dependent static configurations (e.g., Lifting, Swing-Bench) and motion-dependent temporal patterns (e.g., Run-Side, SkateBoarding-Front), providing an ideal testbed for comparing spatial versus temporal modeling approaches. The inclusion of viewpoint variations within the same action type (such as multiple golf swing perspectives) enables systematic analysis of viewpoint invariance capabilities across different representation learning paradigms.

\subsubsection{Video Preprocessing and Standardization}
All video sequences undergo standardized preprocessing to ensure fair comparison between representation learning approaches. Videos are resized to 256×256 resolution while maintaining aspect ratio through center cropping. Frame rates are normalized to 30 fps through temporal interpolation to ensure consistent temporal sampling across sequences. Color space normalization follows ImageNet preprocessing standards with mean subtraction and standard deviation scaling applied channel-wise.

To address the computational constraints inherent in comparative analysis while preserving essential temporal information, we implement a systematic keyframe extraction methodology rather than processing complete video sequences. This approach enables controlled comparison between frame-based and sequence-based architectures while maintaining the temporal richness necessary for meaningful action analysis.

\subsection{Keyframe Extraction Methodology}

\subsubsection{DWT-VGG16-LDA Framework}
Our keyframe selection employs the DWT-VGG16-LDA framework introduced by \citet{kodathala2025sv33bsportsvideounderstanding}, which systematically identifies representative frames corresponding to critical biomechanical phases in athletic movements. This approach addresses the fundamental challenge of temporal sampling in video analysis by combining wavelet-based motion analysis with deep feature extraction and discriminant analysis for optimal frame selection.

The methodology operates through a multi-stage pipeline that processes video content to extract exactly 16 keyframes per sequence, chosen to balance temporal coverage with computational efficiency. The Discrete Wavelet Transform component employs Haar wavelets at decomposition level L=2 to compute wavelet approximation coefficients $(A_i, a_i)$ for each frame $F_i$. This configuration provides optimal localization properties in both time and frequency domains, enabling effective detection of rapid biomechanical transitions characteristic of athletic movements.

Motion analysis proceeds through computation of wavelet approximation differences $D_i = A_{i+1} - A_i$ between consecutive frames, creating robust motion representations that remain stable under varying lighting conditions. These motion maps undergo conversion to 3-channel RGB format to enable processing through convolutional architectures designed for natural images.

\subsubsection{Dual-Path Feature Extraction}
The framework implements parallel feature extraction pathways to capture complementary spatial and temporal information. The appearance pathway processes original RGB frames through a pre-trained VGG-16 model \citep{simonyan2014very} to extract static visual features $f_{a,i}$ that encode posture, positioning, and spatial configuration. Simultaneously, the motion pathway processes wavelet difference maps through the same VGG-16 architecture to extract dynamic features $f_{m,i}$ representing movement patterns and temporal transitions.

Feature fusion combines appearance and motion representations into unified vectors $F = [F_a, F_m]$, creating a high-dimensional space that simultaneously encodes spatial appearance and temporal dynamics. K-means clustering with $K=16$ applied to this fused feature space identifies natural groupings of similar frames, while Linear Discriminant Analysis provides dimensionality reduction that maximizes separation between identified clusters while preserving intra-cluster coherence.

The final keyframe selection identifies the frame within each cluster whose feature representation exhibits minimum Euclidean distance to the cluster centroid in the LDA-transformed space. This approach ensures that selected keyframes represent distinct phases of the athletic action while collectively providing comprehensive temporal coverage of the complete movement sequence.

\subsection{Feature Extraction Protocols}

\subsubsection{DINOv3 Feature Extraction}
DINOv3 features are extracted using the pre-trained facebook/dinov3-vitl16-pretrain-lvd1689m model \citep{simeoni2025dinov3}, which employs a Vision Transformer Large architecture trained on large-scale image collections through self-supervised distillation. The model processes individual keyframes independently, generating 768-dimensional dense visual representations that capture rich spatial semantics, object relationships, and visual attributes within single frames.

Our extraction protocol processes each of the 16 selected keyframes through the DINOv3 encoder independently, producing frame-level feature vectors $v_i \in \mathbb{R}^{768}$ for $i = 1, ..., 16$. These individual frame representations are subsequently aggregated through temporal pooling to create sequence-level representations. We evaluate multiple aggregation strategies including mean pooling, max pooling, and attention-weighted combinations to determine optimal temporal fusion approaches for frame-based representations.

The independence of frame-level processing in DINOv3 enables parallel computation and maintains computational efficiency, while the rich spatial representations learned through large-scale self-supervised pre-training provide detailed visual understanding. However, this approach relies entirely on post-hoc temporal aggregation to capture sequential dynamics, potentially limiting the model's ability to encode complex temporal dependencies and motion patterns.

\subsubsection{V-JEPA2 Feature Extraction}
V-JEPA2 features utilize the facebook/vjepa2-vitl-fpc64-256 model \citep{assran2025vjepa2}, which implements the Video Joint Embedding Predictive Architecture framework with Vision Transformer Large backbone. Unlike DINOv3's frame-independent processing, V-JEPA2 jointly processes the complete 16-frame sequence to generate spatiotemporal representations that explicitly model temporal dependencies and cross-frame relationships.

The V-JEPA2 model processes input sequences of shape $(16, 3, 256, 256)$ through spatiotemporal tokenization, creating patch-based representations that encode both spatial and temporal information. The self-supervised pre-training employs masked prediction objectives in learned representation space, enabling the model to capture motion dynamics, temporal coherence, and frame relationships without pixel-level reconstruction.

Our extraction protocol processes the complete 16-frame keyframe sequence through the V-JEPA2 encoder, producing sequence-level representations $v_{seq} \in \mathbb{R}^{1024}$ that encode the full spatiotemporal context of the athletic action. The joint processing approach enables the model to capture subtle motion patterns, temporal dependencies, and dynamic interactions that unfold across multiple frames, potentially providing superior representation of complex athletic movements.

\subsection{Evaluation Framework}

\subsubsection{Multi-Dimensional Assessment Protocol}
Our evaluation framework employs a comprehensive multi-dimensional assessment that extends beyond traditional classification accuracy to examine fundamental representational characteristics across multiple dimensions of video understanding performance. This approach provides deeper insights into the relative strengths and limitations of frame-based versus sequence-based representation learning paradigms.

\paragraph{Clustering Performance Analysis}
We assess the quality of learned representations through unsupervised clustering analysis using multiple metrics. Silhouette score analysis evaluates cluster coherence by measuring the ratio of inter-cluster to intra-cluster distances, with higher scores indicating better-separated clusters. Calinski-Harabasz scores assess cluster compactness and separation through the ratio of between-cluster dispersion to within-cluster dispersion. These metrics provide insight into the natural grouping structure induced by different representation learning approaches without relying on supervised classification performance.

\paragraph{Similarity Analysis Framework}
Comprehensive similarity analysis examines both intra-class consistency and inter-class discrimination capabilities through cosine similarity measurements in the learned feature space. Intra-class analysis computes pairwise similarities between samples within the same action category, assessing the model's ability to generate consistent representations for similar athletic movements. Inter-class analysis measures similarities across different action categories, evaluating discrimination capabilities and semantic separation.

The similarity analysis framework generates multiple analytical products: (1) pairwise similarity matrices examining all sample combinations, (2) action-specific consistency measurements within each sports category, (3) cross-action discrimination analysis between different movement types, and (4) viewpoint invariance assessment for actions captured from multiple perspectives (e.g., Golf-Swing-Back, Golf-Swing-Front, Golf-Swing-Side).

\paragraph{Nearest Neighbor Classification}
K-nearest neighbor classification with $k \in \{1, 3, 5\}$ evaluates representation quality through non-parametric classification in the learned feature space. This approach assesses whether semantically similar actions cluster together naturally without requiring trained classifiers, providing insight into the semantic organization of learned representations. Classification accuracy across different values of $k$ reveals the local neighborhood structure and consistency of representations.

\paragraph{Dimensionality Reduction and Visualization}
We employ multiple dimensionality reduction techniques to visualize learned representations and assess their semantic organization. t-SNE analysis \citep{vandermaaten2008visualizing} with perplexity values adapted to dataset size provides non-linear visualization of local neighborhood structures. UMAP analysis \citep{mcinnes2018umap} offers complementary global structure preservation for understanding large-scale organization of the feature space. These visualizations enable qualitative assessment of semantic clustering and reveal potential advantages of different representation learning approaches.\\

\section{Results and Analysis}

This section presents a comprehensive comparative analysis of DINOv3 and V-JEPA2 feature representations across multiple dimensions of video understanding performance. Our evaluation encompasses clustering quality assessment, similarity analysis, classification performance, and visualization-based insights to characterize the fundamental trade-offs between frame-based and sequence-based self-supervised learning approaches.

\subsection{Dataset Statistics and Processing}

The experimental evaluation processed 140 video sequences from the UCF Sports dataset, with DINOv3 successfully extracting features from 133 samples (94.3\% success rate) and V-JEPA2 processing all 140 samples. The slight difference in sample counts results from DINOv3's frame-by-frame processing encountering occasional extraction failures on corrupted or malformed video frames, while V-JEPA2's joint sequence processing demonstrates greater robustness to individual frame irregularities.

Both approaches generate high-dimensional feature representations: DINOv3 produces 768-dimensional vectors through spatial feature aggregation, while V-JEPA2 generates 1024-dimensional spatiotemporal representations. The extracted features span 9 distinct action categories with varying sample distributions, enabling comprehensive analysis across diverse athletic movements ranging from static poses (Lifting) to dynamic sequences (SkateBoarding, Run).

\subsection{Clustering Performance Analysis}

Table \ref{tab:clustering_metrics} presents the clustering quality metrics for both representation learning approaches, revealing fundamental differences in their ability to organize action categories in feature space.

\begin{table}
	\caption{Clustering Performance Metrics Comparison}
	\centering
	\begin{tabular}{lcc}
		\toprule
		Metric & DINOv3 & V-JEPA2 \\
		\midrule
		Silhouette Score & \textbf{0.310} & 0.206 \\
		Calinski-Harabasz Index & \textbf{12.51} & 8.76 \\
		Feature Dimension & 768 & 1024 \\
		\bottomrule
	\end{tabular}
	\label{tab:clustering_metrics}
\end{table}

DINOv3 demonstrates superior clustering performance with a Silhouette score of 0.310 compared to V-JEPA2's 0.206, indicating significantly better-separated action clusters in the learned feature space. The higher Calinski-Harabasz index (12.51 vs 8.76) further confirms DINOv3's ability to create more compact intra-class clusters with greater inter-class separation. These results suggest that frame-based spatial processing excels at creating discriminative representations that naturally group similar actions while maintaining clear boundaries between different movement categories.

\subsection{Classification Performance Evaluation}

K-nearest neighbor classification results, presented in Table \ref{tab:knn_performance}, provide insight into the practical utility of learned representations for downstream recognition tasks.

\begin{table}
	\caption{K-Nearest Neighbor Classification Accuracy}
	\centering
	\begin{tabular}{lccc}
		\toprule
		Method & k=1 & k=3 & k=5 \\
		\midrule
		DINOv3 & \textbf{0.895} & \textbf{0.880} & \textbf{0.857} \\
		V-JEPA2 & 0.879 & 0.871 & 0.843 \\
		\bottomrule
	\end{tabular}
	\label{tab:knn_performance}
\end{table}

DINOv3 achieves consistently higher classification accuracies across all neighborhood sizes, with particularly strong performance in k=1 classification (89.5\% vs 87.9\%). The performance gap narrows slightly as k increases, suggesting that both approaches benefit from larger neighborhood averaging, but DINOv3 maintains its advantage throughout. These results align with the clustering analysis, confirming that DINOv3's superior cluster separation translates directly into improved classification performance.

\subsection{Similarity Analysis}

\subsubsection{Intra-Class Consistency}

\begin{figure}
	\includegraphics[width=0.75\textwidth]{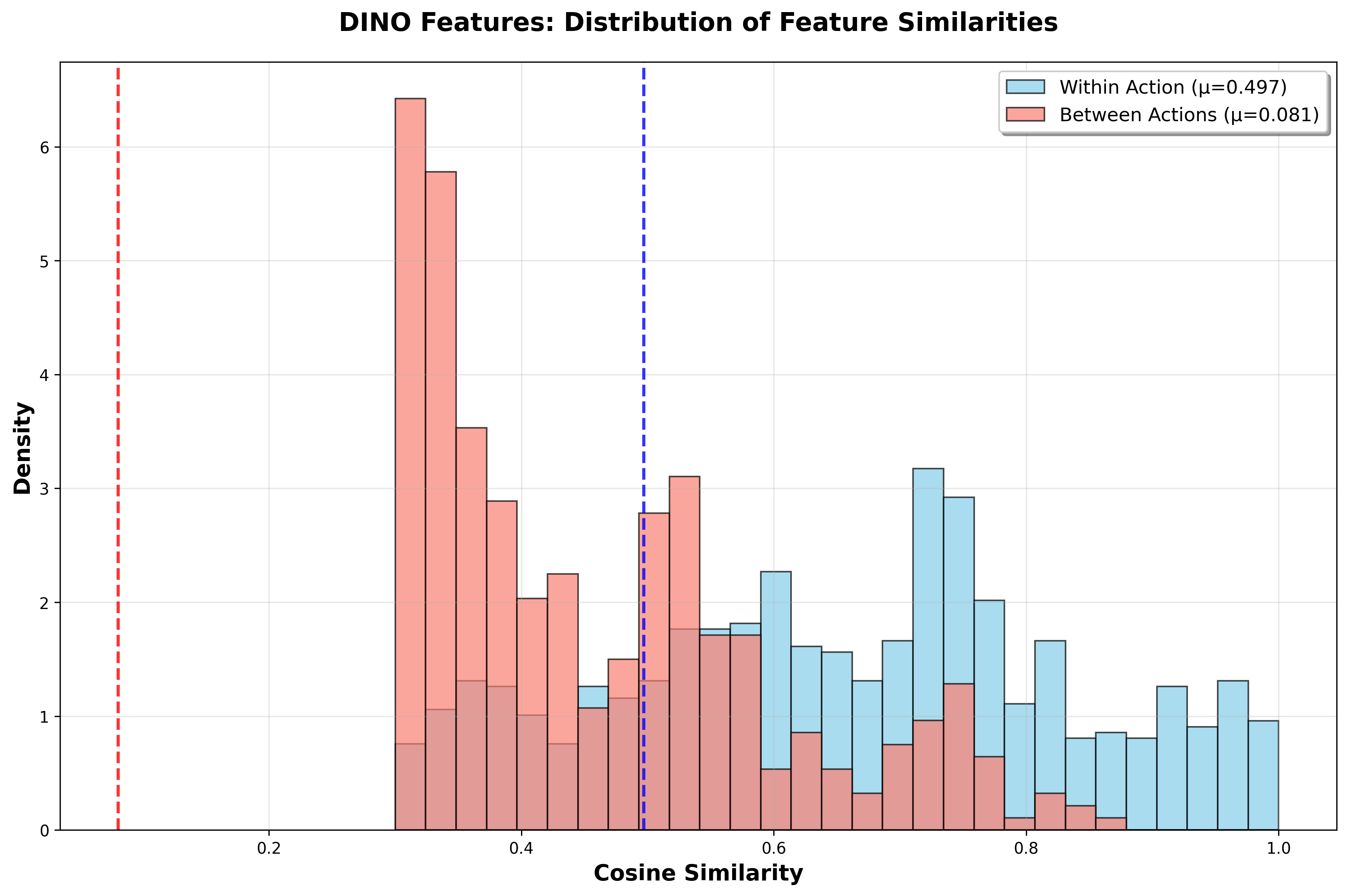}
	\includegraphics[width=0.75\textwidth]{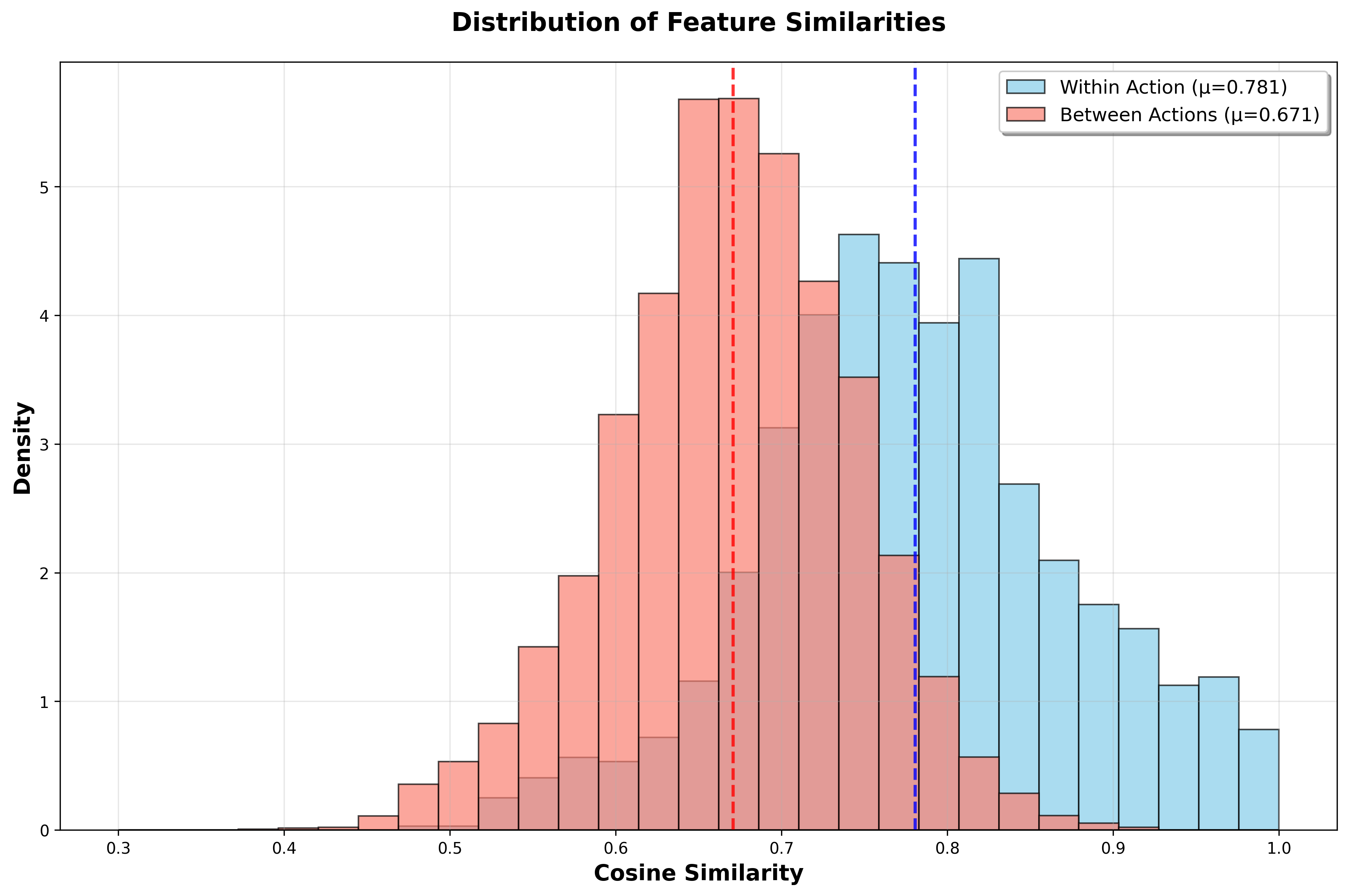}
	\centering
	\caption{Distribution of Feature Similarities - DINOv3 (top) and V-JEPA2 (bottom)}
	\label{fig:similarity_distributions}
\end{figure}

Figure \ref{fig:similarity_distributions} reveals striking differences in similarity distribution patterns between the two approaches. DINOv3 exhibits a clear bimodal distribution with distinct peaks for intra-class ($\mu$=0.497) and inter-class ($\mu$=0.081) similarities, demonstrating exceptional discrimination capability with a 6.16× separation ratio. V-JEPA2 shows more overlapping distributions (intra-class $\mu$=0.781, inter-class $\mu$=0.671), indicating higher overall similarity values but reduced discriminative power.

\begin{figure}
	\centering
	\includegraphics[width=0.75\textwidth]{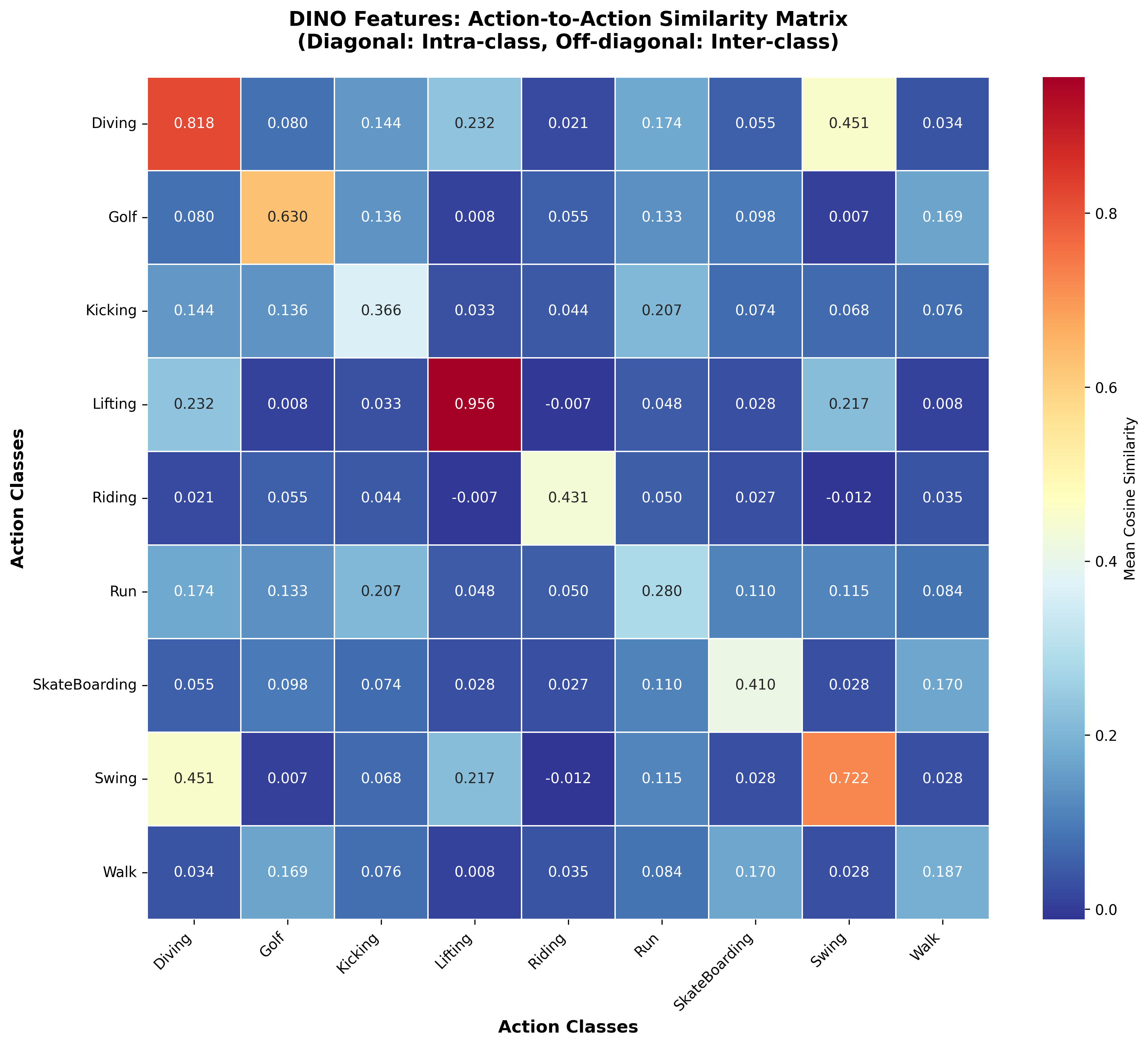}
	\caption{DINOv3 Features: Action-to-Action Similarity Matrix}
	\label{fig:dino_action_matrix}
\end{figure}

\begin{figure}
	\centering
	\includegraphics[width=0.75\textwidth]{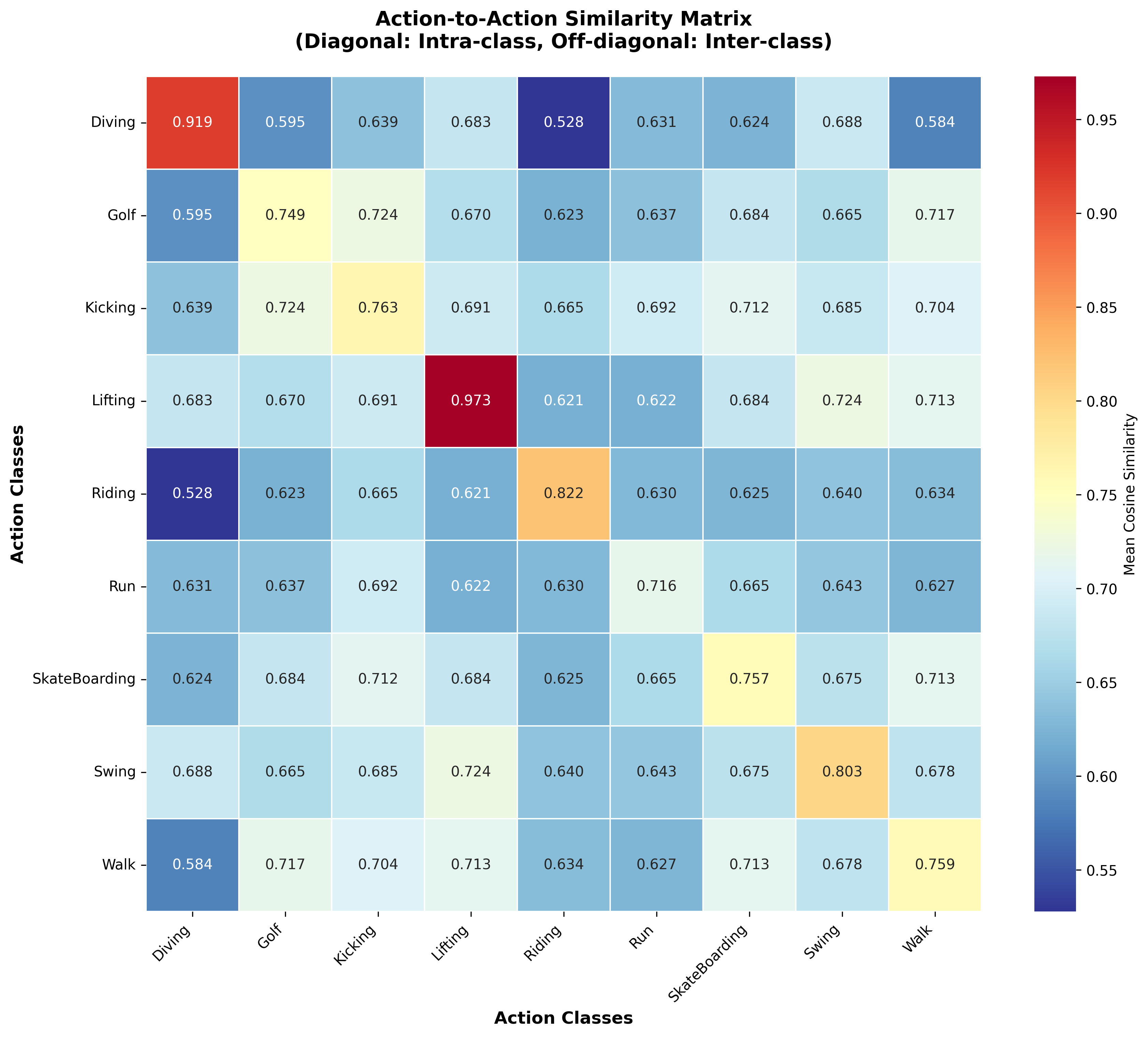}
	\caption{V-JEPA2 Features: Action-to-Action Similarity Matrix}
	\label{fig:jepa2_action_matrix}
\end{figure}

The action-to-action similarity matrices (Figures \ref{fig:dino_action_matrix} and \ref{fig:jepa2_action_matrix}) provide detailed insight into architectural differences. DINOv3 demonstrates exceptional discriminative capability with very low inter-class similarities (typically <0.2) and variable intra-class consistency ranging from 0.187 (Walk) to 0.956 (Lifting). V-JEPA2 exhibits consistently higher similarities across all action pairs, with intra-class similarities ranging from 0.716 (Run) to 0.973 (Lifting) and inter-class similarities between 0.528-0.717.

\subsubsection{Action-Specific Performance Patterns}

\begin{figure}
	\centering
	\includegraphics[width=0.75\textwidth]{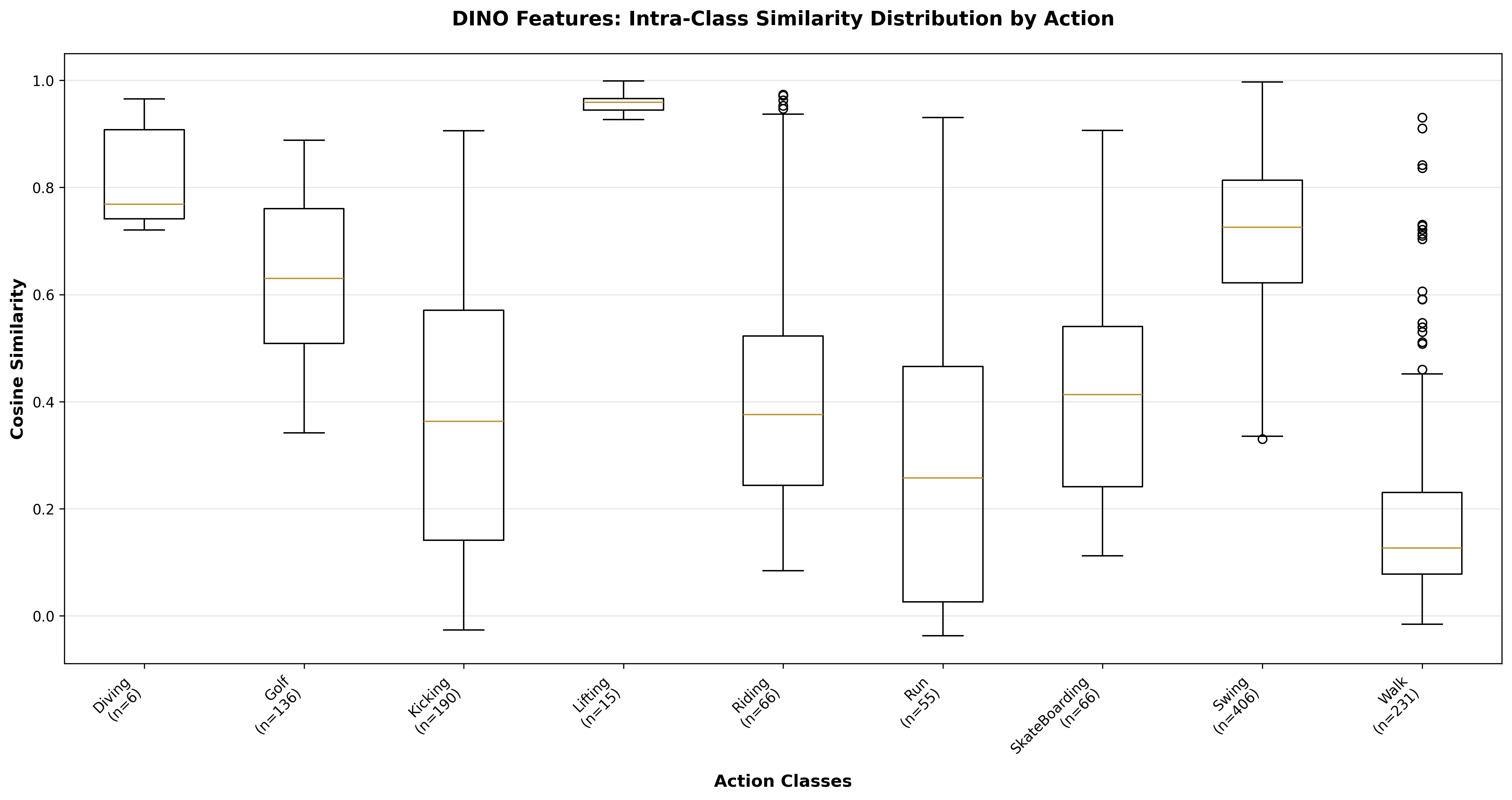}
	\caption{DINOv3 Features: Intra-Class Similarity Distribution by Action}
	\label{fig:dino_intra_class}
\end{figure}

\begin{figure}
	\centering
	\includegraphics[width=0.75\textwidth]{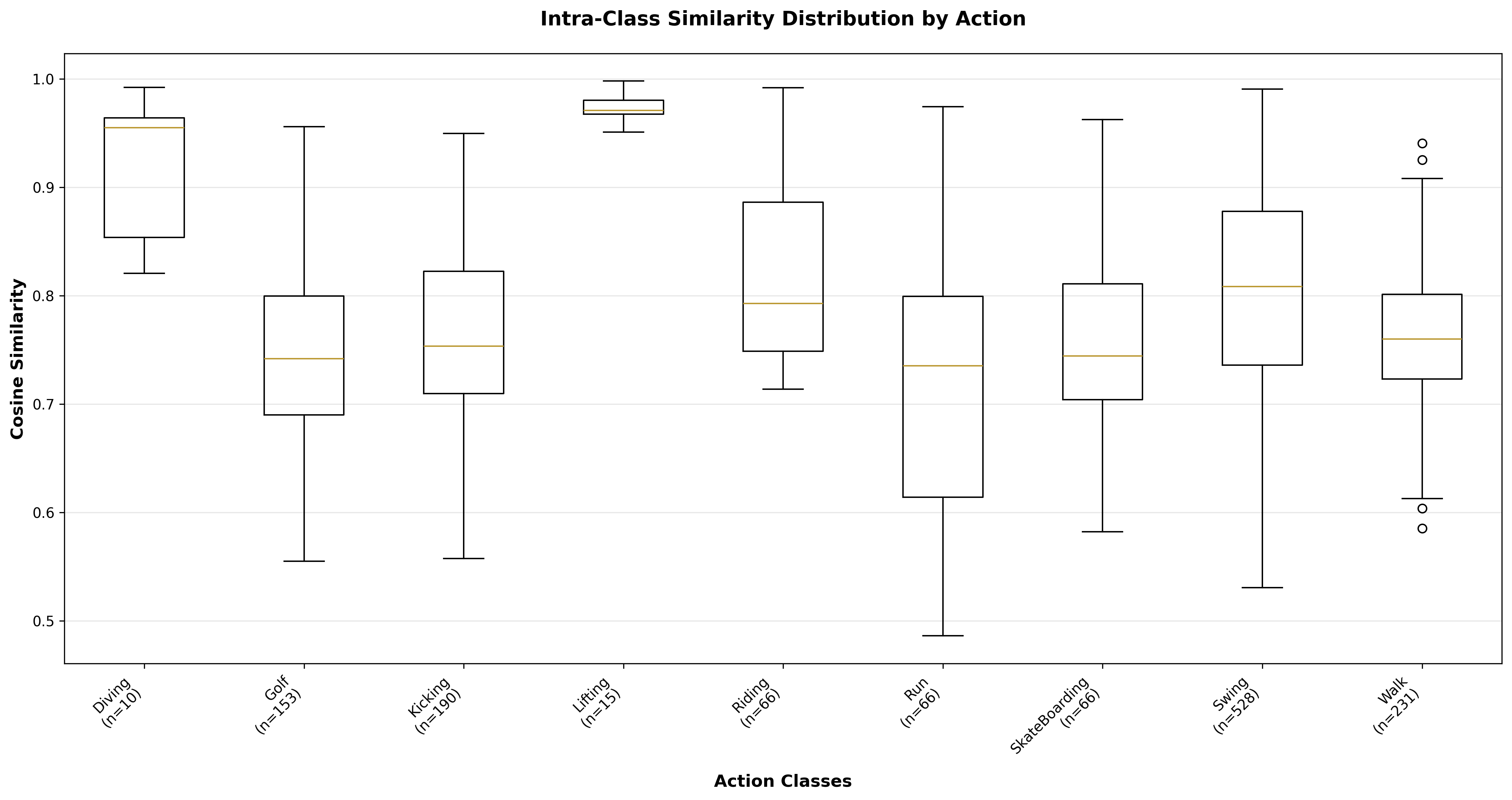}
	\caption{V-JEPA2 Features: Intra-Class Similarity Distribution by Action}
	\label{fig:jepa2_intra_class}
\end{figure}

Figures \ref{fig:dino_intra_class} and \ref{fig:jepa2_intra_class} illustrate per-action intra-class similarity distributions, revealing systematic performance patterns across different movement types.

DINOv3 exhibits dramatic performance variation across action categories, excelling at pose-identifiable actions like Lifting (0.956 ± 0.020) and Diving (0.818 ± 0.111) while struggling with motion-dependent actions such as Walk (0.187 ± 0.177) and Run (0.280 ± 0.267). This pattern reflects the architectural bias toward spatial processing, where static configurations are more easily captured than temporal dynamics.

Conversely, V-JEPA2 demonstrates remarkable consistency across all action types, with intra-class similarities ranging from 0.716 (Run) to 0.973 (Lifting) and substantially lower variance ($\sigma$= 0.094 vs DINOv3's $\sigma$ = 0.288). This consistency suggests that joint temporal processing provides reliable representations across diverse movement patterns, avoiding the dramatic performance degradation observed in frame-based approaches for temporally complex actions.

\subsection{Hierarchical Clustering Analysis}

\begin{figure}
	\centering
		\includegraphics[width=0.75\textwidth]{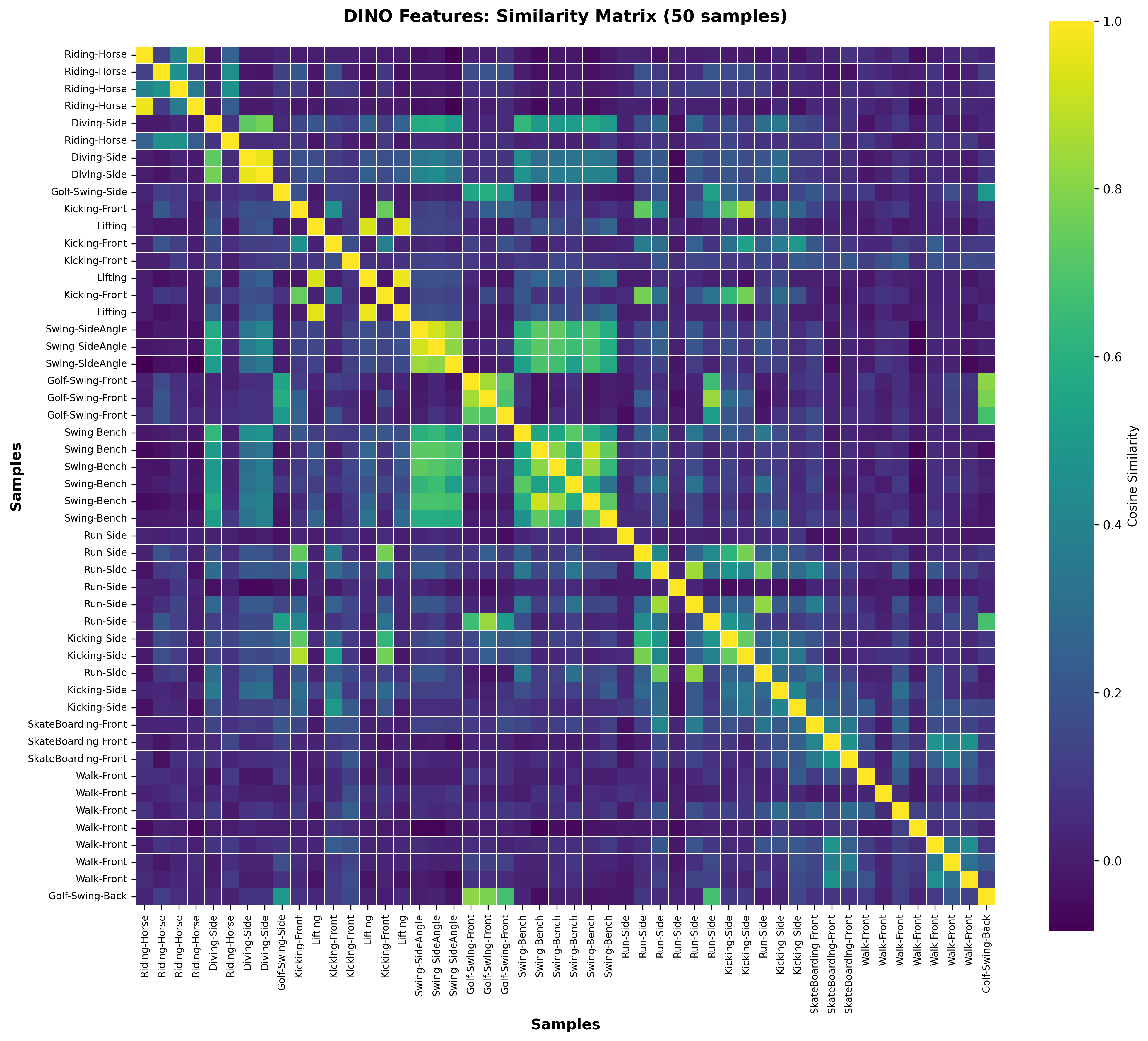}
	\caption{DINOv3 Features: Hierarchical Clustering (50 samples)}
	\label{fig:dino_hierarchical}
\end{figure}

\begin{figure}
	\centering
	\includegraphics[width=0.75\textwidth]{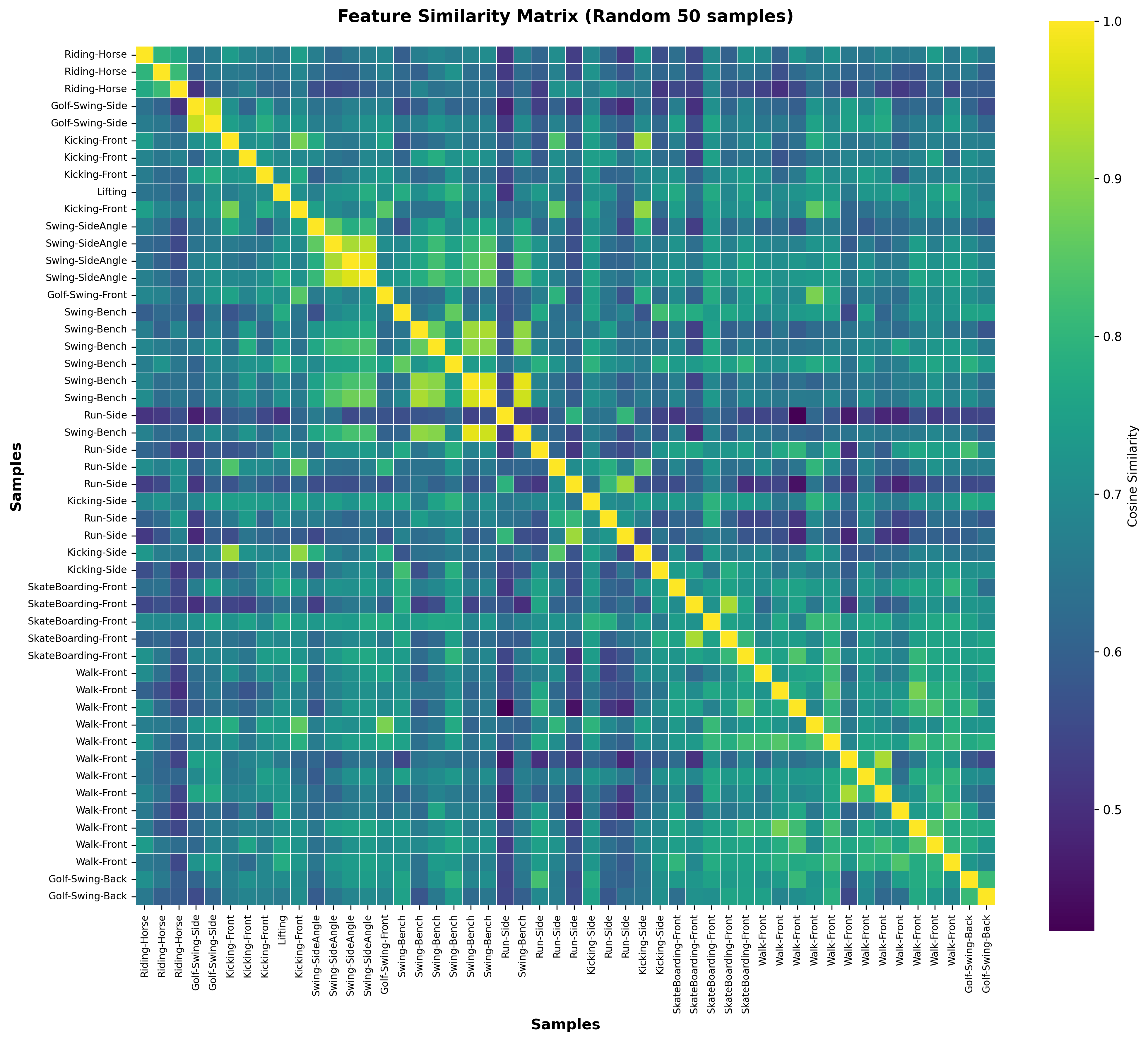}
		\caption{VJEPA2 Features: Hierarchical Clustering (50 samples)}
	\label{fig:jepa2_hierarchical}
\end{figure}

Hierarchical clustering dendrograms (Figures \ref{fig:dino_hierarchical} and \ref{fig:jepa2_hierarchical}) provide qualitative insight into the learned semantic organization of action categories.

DINOv3 clustering exhibits clear action-specific groupings with distinct separation between different movement categories, particularly evident in the tight clustering of similar swing-based actions. The dendrogram reveals a hierarchical organization that reflects intuitive semantic relationships between sports activities. V-JEPA2 shows more gradual clustering with less pronounced category separation, suggesting that temporal features create smoother transitions between related actions but potentially sacrifice discriminative clarity.

\subsection{Dimensionality Reduction and Visualization}

\subsubsection{t-SNE Analysis}

\begin{figure}
	\centering
	\includegraphics[width=0.75\textwidth]{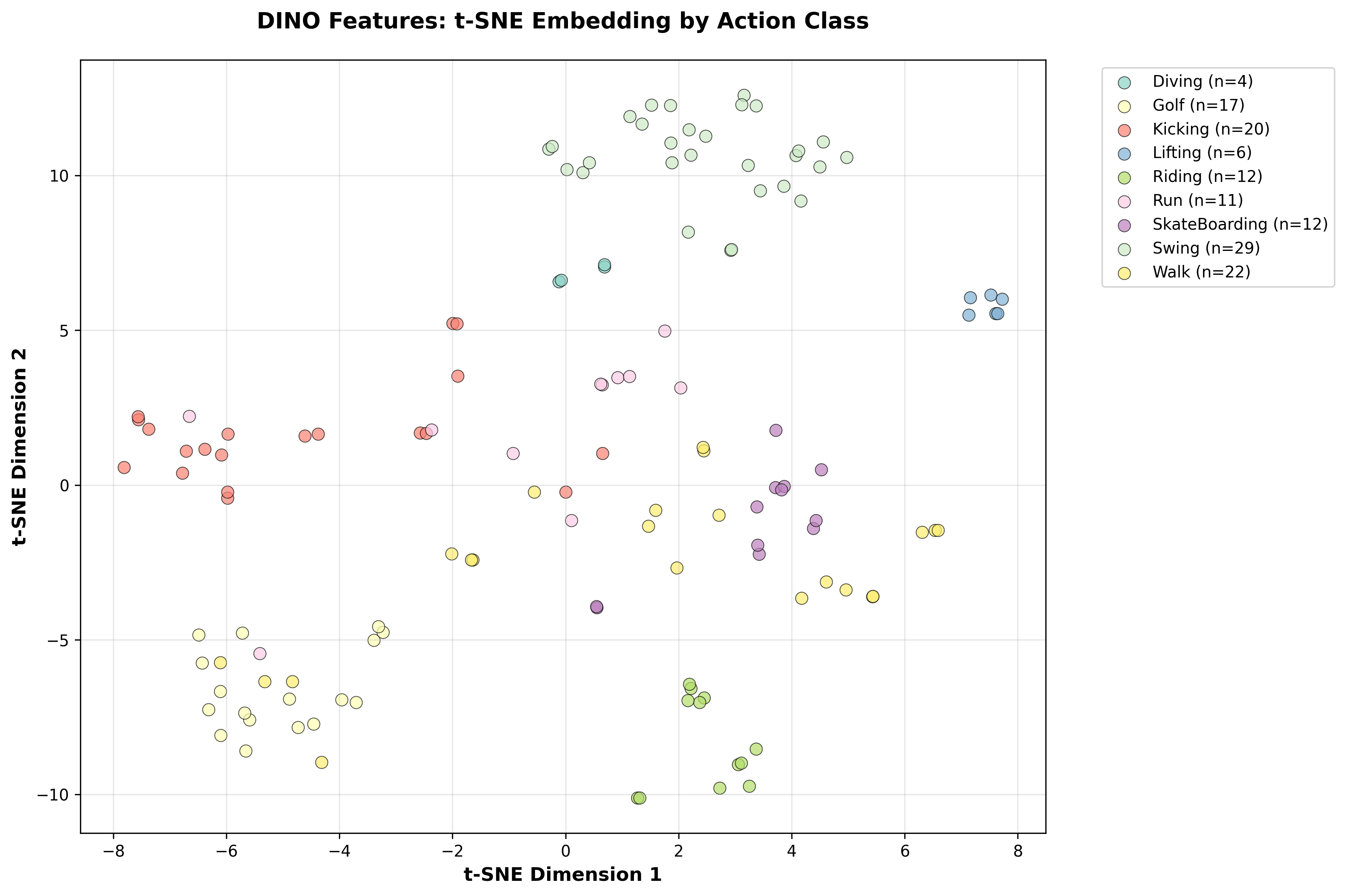}
	\caption{DINOv3 Features: t-SNE Embedding by Action Class}
	\label{fig:dino_tsne}
\end{figure}

\begin{figure}
	\centering
	\includegraphics[width=0.75\textwidth]{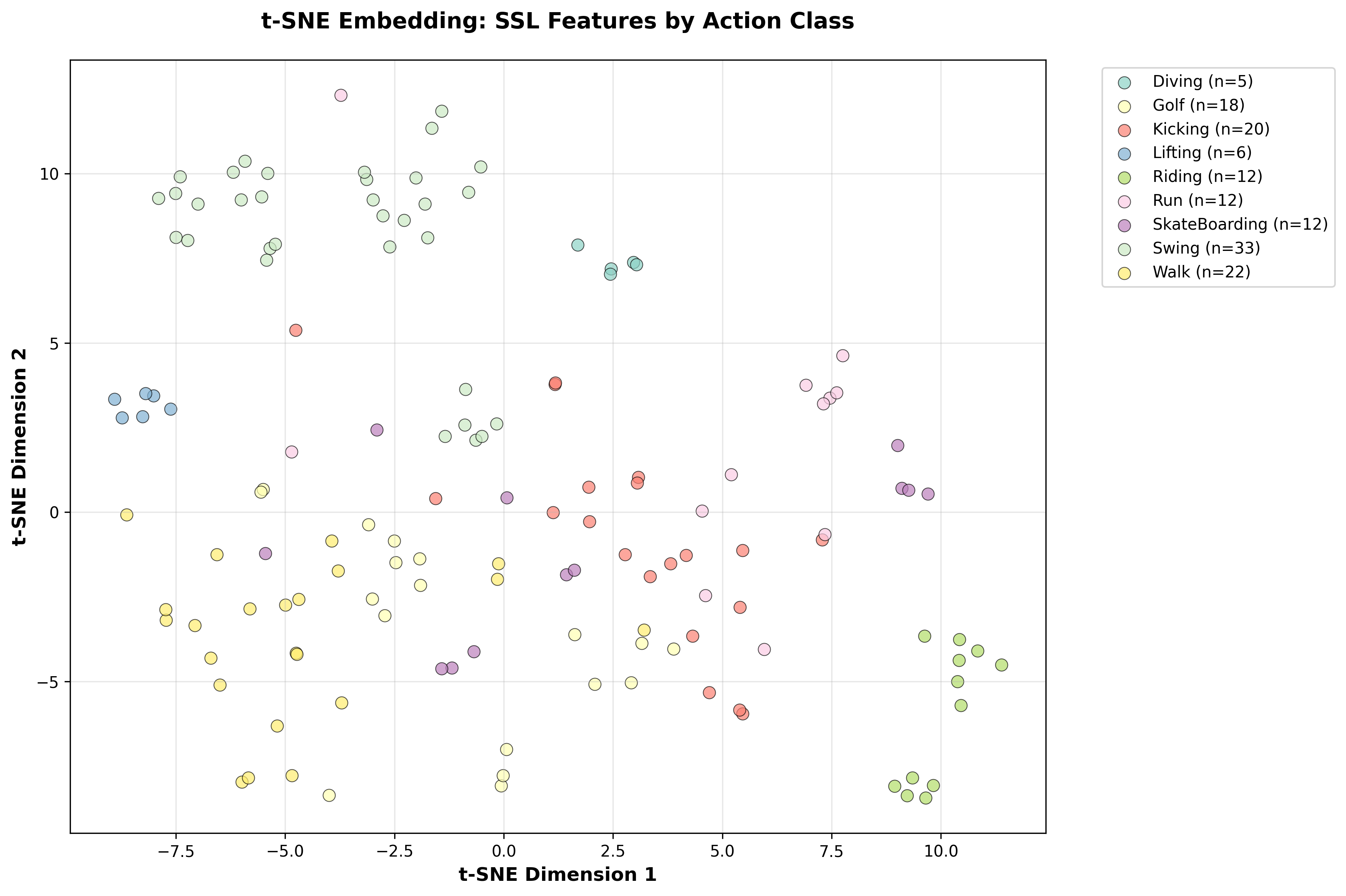}
	\caption{V-JEPA2 Features: t-SNE Embedding by Action Class}
	\label{fig:jepa2_tsne}
\end{figure}

T-SNE embeddings (Figures \ref{fig:dino_tsne} and \ref{fig:jepa2_tsne}) reveal fundamental differences in feature space organization between the two approaches.

DINOv3 t-SNE visualization displays well-separated action clusters with minimal overlap, particularly for pose-dependent actions like Lifting, Diving, and Swing categories. The clear separation confirms the quantitative clustering results and demonstrates the model's ability to learn discriminative spatial features. However, motion-intensive actions like Walk and Run show more dispersed representations, reflecting the temporal modeling limitations of frame-based approaches.

V-JEPA2 t-SNE embeddings exhibit more compact overall organization with smoother transitions between related action categories. While individual clusters are less distinctly separated, the visualization reveals a more uniform distribution of samples within each action category, supporting the consistency findings from similarity analysis.

\subsubsection{UMAP Analysis}

\begin{figure}
	\centering
	\includegraphics[width=0.75\textwidth]{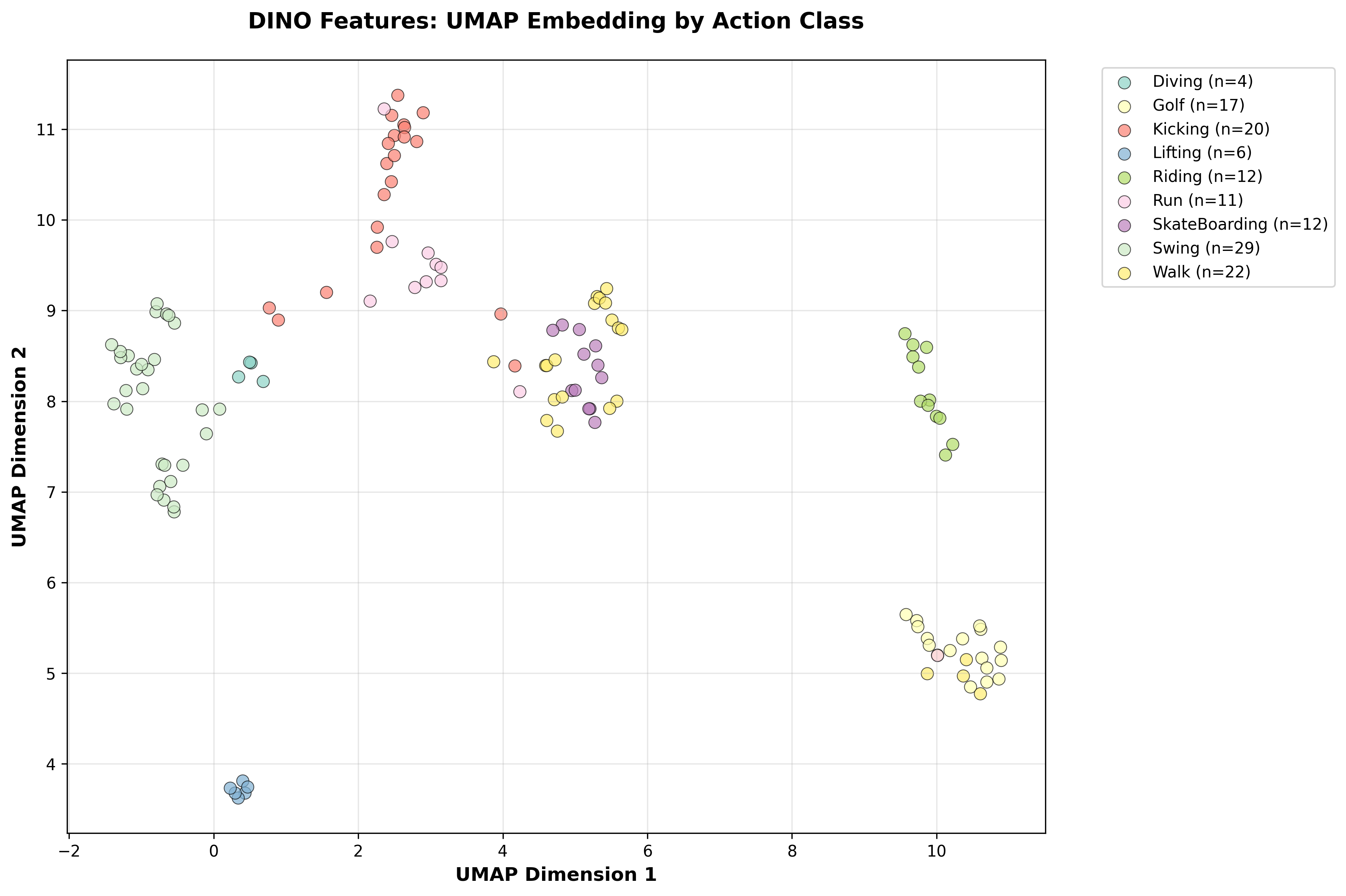}
	\caption{DINOv3 Features: UMAP Embedding by Action Class}
	\label{fig:dino_umap}
\end{figure}

\begin{figure}
	\centering
	\includegraphics[width=0.75\textwidth]{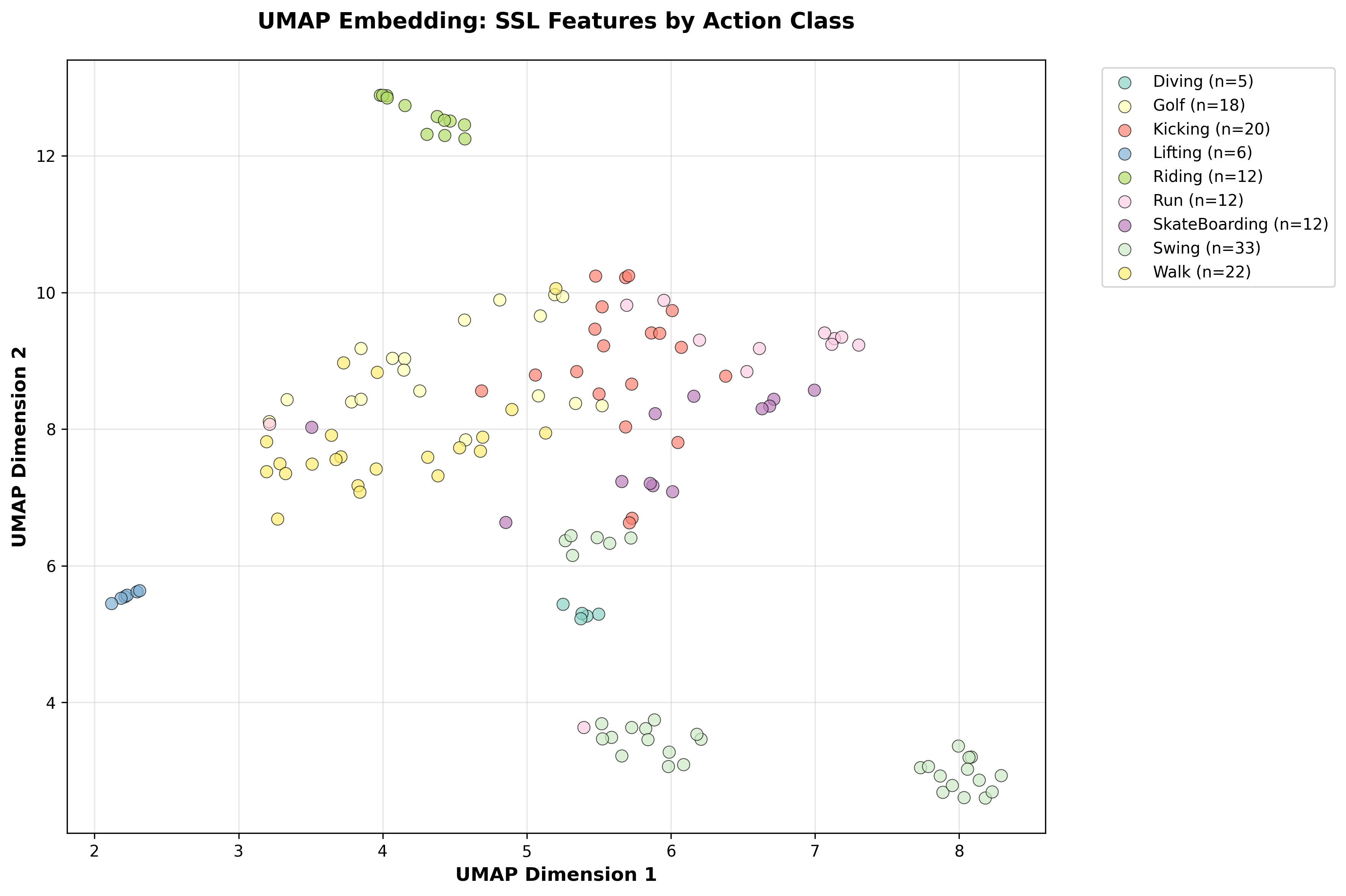}
	\caption{V-JEPA2 Features: UMAP Embedding by Action Class}
	\label{fig:jepa2_umap}
\end{figure}

UMAP embeddings (Figures \ref{fig:dino_umap} and \ref{fig:jepa2_umap}) provide complementary insight into global feature space structure preservation.

DINOv3 UMAP visualization reinforces the t-SNE findings with distinct action clusters and clear inter-class boundaries. The preservation of global structure reveals semantic relationships between actions, with related activities (different swing types, various movement patterns) positioned in nearby regions. V-JEPA2 UMAP demonstrates more overlapping clusters with preserved local neighborhood structures, indicating that temporal modeling creates features that capture subtle motion relationships across action boundaries.

\subsection{Performance Variance and Reliability Analysis}

A critical finding emerges from the variance analysis across action categories. DINOv3 exhibits high performance variance ($\sigma$ = 0.288) with exceptional performance on pose-identifiable actions but significant degradation on temporally complex movements. This creates a reliability challenge where performance is unpredictable based on action type.

V-JEPA2's substantially lower performance variance ($\sigma$ = 0.094) indicates consistent reliability across diverse action vocabularies. While the peak performance may not match DINOv3's best cases, the predictable performance across all action types suggests greater suitability for production deployment scenarios requiring stable behavior.

The consistency analysis reveals that V-JEPA2's joint temporal modeling provides crucial advantages for comprehensive video understanding tasks, particularly when dealing with diverse action vocabularies that encompass both spatial configurations and temporal dynamics. DINOv3's architectural design, while achieving superior discrimination for pose-based recognition, struggles with temporally-dependent actions that require sequential pattern understanding.

\section{Conclusions}
Our comprehensive evaluation demonstrates that the choice between DINOv3 and V-JEPA2 for video action analysis depends critically on task-specific requirements and tolerance for performance variability. While DINOv3 achieves marginally superior overall classification accuracy (89.47\% vs 87.86\%) and significantly better clustering performance, it exhibits substantial inconsistency across action types, with performance ranging from 0.19 to 0.96 in intra-class consistency metrics. Conversely, V-JEPA2 provides reliable performance across all evaluated action categories (0.72-0.97 range) with three times lower variance, making it more suitable for production deployment scenarios requiring predictable behavior. The architectural differences manifest clearly in action-specific performance patterns, where DINOv3's spatial processing excels at pose-based recognition tasks but struggles with temporally-dependent actions that require sequential pattern understanding. This study establishes that temporal modeling capabilities, as embodied in V-JEPA2's architecture, provide crucial advantages for comprehensive video understanding tasks, particularly when dealing with diverse action vocabularies that encompass both spatial configurations and temporal dynamics. Future work should investigate hybrid architectures that combine DINOv3's discriminative spatial processing with V-JEPA2's temporal consistency to achieve both high performance and reliability across varied action recognition scenarios.

\bibliographystyle{unsrtnat}
\bibliography{references}  

\begin{thebibliography}{67}
\providecommand{\natexlab}[1]{#1}
\providecommand{\url}[1]{\texttt{#1}}
\expandafter\ifx\csname urlstyle\endcsname\relax
  \providecommand{\doi}[1]{doi: #1}\else
  \providecommand{\doi}{doi: \begingroup \urlstyle{rm}\Url}\fi

\bibitem[Aggarwal and Ryoo(2011)]{aggarwal2011human}
Jake~K Aggarwal and Michael~S Ryoo.
\newblock Human activity analysis: A review.
\newblock \emph{ACM Computing Surveys}, 43\penalty0 (3):\penalty0 1--43, 2011.
\newblock \doi{10.1145/1922649.1922653}.

\bibitem[Moeslund et~al.(2006)Moeslund, Hilton, and
  Kr{\"u}ger]{moeslund2006survey}
Thomas~B Moeslund, Adrian Hilton, and Volker Kr{\"u}ger.
\newblock A survey of advances in vision-based human motion capture and
  analysis.
\newblock \emph{Computer Vision and Image Understanding}, 104\penalty0
  (2-3):\penalty0 90--126, 2006.
\newblock \doi{10.1016/j.cviu.2006.08.002}.

\bibitem[Schiappa et~al.(2023)Schiappa, Rawat, and Shah]{schiappa2023self}
Madeline~C Schiappa, Yogesh~S Rawat, and Mubarak Shah.
\newblock Self-supervised learning for videos: A survey.
\newblock \emph{ACM Computing Surveys}, 55\penalty0 (13s):\penalty0 1--37,
  2023.
\newblock \doi{10.1145/3577925}.

\bibitem[Wang et~al.(2016)Wang, Xiong, Wang, Qiao, Lin, Tang, and
  Van~Gool]{wang2016temporal}
Limin Wang, Yuanjun Xiong, Zhe Wang, Yu~Qiao, Dahua Lin, Xiaoou Tang, and Luc
  Van~Gool.
\newblock Temporal segment networks: Towards good practices for deep action
  recognition.
\newblock In \emph{European Conference on Computer Vision}, pages 20--36.
  Springer, 2016.
\newblock \doi{10.1007/978-3-319-46484-8_2}.

\bibitem[Krizhevsky et~al.(2012)Krizhevsky, Sutskever, and
  Hinton]{krizhevsky2012imagenet}
Alex Krizhevsky, Ilya Sutskever, and Geoffrey~E Hinton.
\newblock Imagenet classification with deep convolutional neural networks.
\newblock In \emph{Advances in Neural Information Processing Systems}, pages
  1097--1105. NIPS, 2012.

\bibitem[Tran et~al.(2015)Tran, Bourdev, Fergus, Torresani, and
  Paluri]{tran2015learning}
Du~Tran, Lubomir Bourdev, Rob Fergus, Lorenzo Torresani, and Manohar Paluri.
\newblock Learning spatiotemporal features with 3d convolutional networks.
\newblock In \emph{Proceedings of the IEEE International Conference on Computer
  Vision}, pages 4489--4497, 2015.
\newblock \doi{10.1109/ICCV.2015.510}.

\bibitem[Carreira and Zisserman(2017)]{carreira2017quo}
Joao Carreira and Andrew Zisserman.
\newblock Quo vadis, action recognition? a new model and the kinetics dataset.
\newblock In \emph{Proceedings of the IEEE Conference on Computer Vision and
  Pattern Recognition}, pages 6299--6308, 2017.
\newblock \doi{10.1109/CVPR.2017.502}.

\bibitem[{Meta AI}(2024)]{metaai2024dinov3}
{Meta AI}.
\newblock Dinov3: Self-supervised learning for vision at unprecedented scale.
\newblock Meta AI Research Blog, 2024.
\newblock URL
  \url{https://ai.meta.com/blog/dinov3-self-supervised-learning-vision/}.

\bibitem[Assran et~al.(2025)]{assran2025vjepa2}
Mahmoud Assran et~al.
\newblock V-jepa 2: Self-supervised video models enable understanding,
  prediction and planning.
\newblock \emph{arXiv preprint arXiv:2506.09985}, 2025.

\bibitem[LeCun(2022)]{lecun2022path}
Yann LeCun.
\newblock A path towards autonomous machine intelligence.
\newblock \emph{OpenReview}, 2022.
\newblock URL \url{https://openreview.net/forum?id=BZ5a1r-kVsf}.

\bibitem[Dosovitskiy et~al.(2021)Dosovitskiy, Beyer, Kolesnikov, Weissenborn,
  Zhai, Unterthiner, Dehghani, Minderer, Heigold, Gelly,
  et~al.]{dosovitskiy2021image}
Alexey Dosovitskiy, Lucas Beyer, Alexander Kolesnikov, Dirk Weissenborn,
  Xiaohua Zhai, Thomas Unterthiner, Mostafa Dehghani, Matthias Minderer, Georg
  Heigold, Sylvain Gelly, et~al.
\newblock An image is worth 16x16 words: Transformers for image recognition at
  scale.
\newblock In \emph{International Conference on Learning Representations}, 2021.

\bibitem[Caron et~al.(2021)Caron, Touvron, Misra, J{\'e}gou, Mairal,
  Bojanowski, and Joulin]{caron2021emerging}
Mathilde Caron, Hugo Touvron, Ishan Misra, Herv{\'e} J{\'e}gou, Julien Mairal,
  Piotr Bojanowski, and Armand Joulin.
\newblock Emerging properties in self-supervised vision transformers.
\newblock In \emph{Proceedings of the IEEE/CVF International Conference on
  Computer Vision}, pages 9650--9660, 2021.
\newblock \doi{10.1109/ICCV.2021.00951}.

\bibitem[Soomro and Zamir(2014)]{soomro2014action}
Khurram Soomro and Amir~R Zamir.
\newblock Action recognition in realistic sports videos.
\newblock In \emph{Computer Vision in Sports}. Springer International
  Publishing, 2014.
\newblock URL \url{https://www.crcv.ucf.edu/data/UCF_Sports_Action.php}.

\bibitem[Donahue et~al.(2015)Donahue, Anne~Hendricks, Guadarrama, Rohrbach,
  Venugopalan, Saenko, and Darrell]{donahue2015long}
Jeffrey Donahue, Lisa Anne~Hendricks, Sergio Guadarrama, Marcus Rohrbach,
  Subhashini Venugopalan, Kate Saenko, and Trevor Darrell.
\newblock Long-term recurrent convolutional networks for visual recognition and
  description.
\newblock In \emph{Proceedings of the IEEE Conference on Computer Vision and
  Pattern Recognition}, pages 2625--2634, 2015.
\newblock \doi{10.1109/CVPR.2015.7298878}.

\bibitem[Pathak et~al.(2017)Pathak, Agrawal, Efros, and
  Darrell]{pathak2017curiosity}
Deepak Pathak, Pulkit Agrawal, Alexei~A Efros, and Trevor Darrell.
\newblock Curiosity-driven exploration by self-supervised prediction.
\newblock In \emph{Proceedings of the IEEE Conference on Computer Vision and
  Pattern Recognition}, pages 2778--2787, 2017.
\newblock \doi{10.1109/CVPR.2017.70}.

\bibitem[Lee et~al.(2017)Lee, Huang, Singh, and Yang]{lee2017unsupervised}
Hsin-Ying Lee, Jia-Bin Huang, Maneesh Singh, and Ming-Hsuan Yang.
\newblock Unsupervised representation learning by sorting sequences.
\newblock In \emph{Proceedings of the IEEE International Conference on Computer
  Vision}, pages 667--676, 2017.
\newblock \doi{10.1109/ICCV.2017.79}.

\bibitem[Kim et~al.(2019)Kim, Cho, and Kweon]{kim2019self}
Dahun Kim, Donghyeon Cho, and In~So Kweon.
\newblock Self-supervised video representation learning with space-time cubic
  puzzles.
\newblock In \emph{Proceedings of the AAAI Conference on Artificial
  Intelligence}, volume~33, pages 8545--8552, 2019.
\newblock \doi{10.1609/aaai.v33i01.33018545}.

\bibitem[Schiappa et~al.(2022)Schiappa, Rawat, and Shah]{schiappa2022self}
Madeline~C Schiappa, Yogesh~S Rawat, and Mubarak Shah.
\newblock Self-supervised learning for videos: A survey.
\newblock \emph{arXiv preprint arXiv:2207.00419}, 2022.

\bibitem[Misra et~al.(2016)Misra, Zitnick, and Hebert]{misra2016shuffle}
Ishan Misra, C~Lawrence Zitnick, and Martial Hebert.
\newblock Shuffle and learn: Unsupervised learning using temporal order
  verification.
\newblock In \emph{European Conference on Computer Vision}, pages 527--544.
  Springer, 2016.
\newblock \doi{10.1007/978-3-319-46448-0_32}.

\bibitem[B{\"u}chler et~al.(2018)B{\"u}chler, Brattoli, and
  Ommer]{buchler2018improving}
Uta B{\"u}chler, Biagio Brattoli, and Bj{\"o}rn Ommer.
\newblock Improving spatiotemporal self-supervision by deep understanding of
  context.
\newblock In \emph{Proceedings of the IEEE Conference on Computer Vision and
  Pattern Recognition}, pages 6582--6591, 2018.
\newblock \doi{10.1109/CVPR.2018.00688}.

\bibitem[Vondrick et~al.(2016)Vondrick, Pirsiavash, and
  Torralba]{vondrick2016generating}
Carl Vondrick, Hamed Pirsiavash, and Antonio Torralba.
\newblock Generating videos with scene dynamics.
\newblock In \emph{Advances in Neural Information Processing Systems}, pages
  613--621. NIPS, 2016.

\bibitem[Srivastava et~al.(2015)Srivastava, Mansimov, and
  Salakhudinov]{srivastava2015unsupervised}
Nitish Srivastava, Elman Mansimov, and Ruslan Salakhudinov.
\newblock Unsupervised learning of video representations using lstms.
\newblock In \emph{International Conference on Machine Learning}, pages
  843--852. PMLR, 2015.

\bibitem[Han et~al.(2020)Han, Xie, and Zisserman]{han2020self}
Tengda Han, Weidi Xie, and Andrew Zisserman.
\newblock Self-supervised co-training for video representation learning.
\newblock In \emph{Advances in Neural Information Processing Systems},
  volume~33, pages 5679--5690. NeurIPS, 2020.

\bibitem[Tian et~al.(2019)Tian, Krishnan, and Isola]{tian2019contrastive}
Yonglong Tian, Dilip Krishnan, and Phillip Isola.
\newblock Contrastive multiview coding.
\newblock \emph{arXiv preprint arXiv:1906.05849}, 2019.

\bibitem[Arandjelovic and Zisserman(2017)]{arandjelovic2017look}
Relja Arandjelovic and Andrew Zisserman.
\newblock Look, listen and learn.
\newblock In \emph{Proceedings of the IEEE International Conference on Computer
  Vision}, pages 609--617, 2017.
\newblock \doi{10.1109/ICCV.2017.73}.

\bibitem[Owens and Efros(2018)]{owens2018audio}
Andrew Owens and Alexei~A Efros.
\newblock Audio-visual scene analysis with self-supervised multisensory
  features.
\newblock In \emph{European Conference on Computer Vision}, pages 631--648.
  Springer, 2018.
\newblock \doi{10.1007/978-3-030-01231-1_39}.

\bibitem[Bardes et~al.(2024)Bardes, Garrido, Ponce, Chen, Rabbat, LeCun,
  Assran, and Ballas]{bardes2024vjepa}
Adrien Bardes, Quentin Garrido, Jean Ponce, Xinlei Chen, Michael Rabbat, Yann
  LeCun, Mahmoud Assran, and Nicolas Ballas.
\newblock V-jepa: Latent video prediction for visual representation learning.
\newblock \emph{OpenReview}, 2024.

\bibitem[Simonyan and Zisserman(2014)]{simonyan2014two}
Karen Simonyan and Andrew Zisserman.
\newblock Two-stream convolutional networks for action recognition in videos.
\newblock In \emph{Advances in Neural Information Processing Systems}, pages
  568--576. NIPS, 2014.

\bibitem[Chen and He(2021)]{chen2021exploring}
Xinlei Chen and Kaiming He.
\newblock Exploring simple siamese representation learning.
\newblock In \emph{Proceedings of the IEEE/CVF Conference on Computer Vision
  and Pattern Recognition}, pages 15750--15758, 2021.
\newblock \doi{10.1109/CVPR46437.2021.01549}.

\bibitem[Oquab et~al.(2023)Oquab, Darcet, Moutakanni, Vo, Szafraniec, Khalidov,
  Fernandez, Haziza, Massa, El-Nouby, et~al.]{oquab2023dinov2}
Maxime Oquab, Timoth{\'e}e Darcet, Th{\'e}o Moutakanni, Huy Vo, Marc
  Szafraniec, Vasil Khalidov, Pierre Fernandez, Daniel Haziza, Francisco Massa,
  Alaaeldin El-Nouby, et~al.
\newblock Dinov2: Learning robust visual features without supervision.
\newblock \emph{arXiv preprint arXiv:2304.07193}, 2023.

\bibitem[Bertasius et~al.(2021)Bertasius, Wang, and
  Torresani]{bertasius2021space}
Gedas Bertasius, Heng Wang, and Lorenzo Torresani.
\newblock Is space-time attention all you need for video understanding?
\newblock In \emph{International Conference on Machine Learning}, pages
  813--824. PMLR, 2021.

\bibitem[Arnab et~al.(2021)Arnab, Dehghani, Heigold, Sun, Lu{\v{c}}i{\'c}, and
  Schmid]{arnab2021vivit}
Anurag Arnab, Mostafa Dehghani, Georg Heigold, Chen Sun, Mario Lu{\v{c}}i{\'c},
  and Cordelia Schmid.
\newblock Vivit: A video vision transformer.
\newblock In \emph{Proceedings of the IEEE/CVF International Conference on
  Computer Vision}, pages 6836--6846, 2021.
\newblock \doi{10.1109/ICCV48922.2021.00676}.

\bibitem[Feichtenhofer et~al.(2016)Feichtenhofer, Pinz, and
  Wildes]{feichtenhofer2016spatiotemporal}
Christoph Feichtenhofer, Axel Pinz, and Richard~P Wildes.
\newblock Spatiotemporal residual networks for video action recognition.
\newblock In \emph{Advances in Neural Information Processing Systems}, pages
  3468--3476. NIPS, 2016.

\bibitem[Ben-Yosef et~al.(2021)Ben-Yosef, Kreiman, and
  Ullman]{benyosef2021minimal}
Guy Ben-Yosef, Leila Kreiman, and Shimon Ullman.
\newblock Minimal videos: Trade-off between spatial and temporal information in
  human and machine vision.
\newblock \emph{Cognition}, 206:\penalty0 104460, 2021.
\newblock \doi{10.1016/j.cognition.2020.104460}.

\bibitem[Huang et~al.(2018)Huang, Fei-Fei, and Niebles]{huang2018analyzing}
De-An Huang, Li~Fei-Fei, and Juan~Carlos Niebles.
\newblock Analyzing temporal information in video understanding models.
\newblock In \emph{Proceedings of the IEEE Conference on Computer Vision and
  Pattern Recognition}, pages 7825--7834, 2018.
\newblock \doi{10.1109/CVPR.2018.00816}.

\bibitem[Duan et~al.(2022)Duan, Zhao, Chen, Lin, and Dai]{duan2022transrank}
Haodong Duan, Yue Zhao, Kai Chen, Dahua Lin, and Bo~Dai.
\newblock Transrank: Self-supervised video representation learning via
  ranking-based transformation recognition.
\newblock In \emph{Proceedings of the IEEE/CVF Conference on Computer Vision
  and Pattern Recognition}, pages 3000--3010, 2022.
\newblock \doi{10.1109/CVPR52688.2022.00302}.

\bibitem[Thomas et~al.(2017)Thomas, Gade, Moeslund, Carr, and
  Hilton]{thomas2017computer}
Graham Thomas, Rikke Gade, Thomas~B Moeslund, Peter Carr, and Adrian Hilton.
\newblock Computer vision for sports: Current applications and research topics.
\newblock \emph{Computer Vision and Image Understanding}, 159:\penalty0 3--18,
  2017.
\newblock \doi{10.1016/j.cviu.2017.04.011}.

\bibitem[Cioppa et~al.(2020)Cioppa, Deliege, Giancola, Ghanem,
  Van~Droogenbroeck, Escalante, and Moeslund]{cioppa2020context}
Anthony Cioppa, Adrien Deliege, Silvio Giancola, Bernard Ghanem, Marc
  Van~Droogenbroeck, Hugo~Jair Escalante, and Thomas~B Moeslund.
\newblock A context-aware loss function for action spotting in soccer videos.
\newblock In \emph{Proceedings of the IEEE/CVF Conference on Computer Vision
  and Pattern Recognition}, pages 13126--13136, 2020.
\newblock \doi{10.1109/CVPR42600.2020.01314}.

\bibitem[Shao et~al.(2020)Shao, Zhao, Dai, and Lin]{shao2020finegym}
Dian Shao, Yue Zhao, Bo~Dai, and Dahua Lin.
\newblock Finegym: A hierarchical video dataset for fine-grained action
  understanding.
\newblock In \emph{Proceedings of the IEEE/CVF Conference on Computer Vision
  and Pattern Recognition}, pages 2616--2625, 2020.
\newblock \doi{10.1109/CVPR42600.2020.00269}.

\bibitem[Rodriguez et~al.(2008)Rodriguez, Ahmed, and Shah]{rodriguez2008action}
Mikel~D Rodriguez, Javed Ahmed, and Mubarak Shah.
\newblock Action mach a spatio-temporal maximum average correlation height
  filter for action recognition.
\newblock In \emph{Proceedings of the IEEE Conference on Computer Vision and
  Pattern Recognition}, pages 1--8, 2008.
\newblock \doi{10.1109/CVPR.2008.4587727}.

\bibitem[Soomro et~al.(2012)Soomro, Zamir, and Shah]{soomro2012ucf101}
Khurram Soomro, Amir~Roshan Zamir, and Mubarak Shah.
\newblock Ucf101: A dataset of 101 human action classes from videos in the
  wild.
\newblock \emph{arXiv preprint arXiv:1212.0402}, 2012.

\bibitem[Kuehne et~al.(2011)Kuehne, Jhuang, Garrote, Poggio, and
  Serre]{kuehne2011hmdb}
Hildegard Kuehne, Hueihan Jhuang, Est{\'\i}baliz Garrote, Tomaso Poggio, and
  Thomas Serre.
\newblock Hmdb: A large video database for human motion recognition.
\newblock In \emph{Proceedings of the IEEE International Conference on Computer
  Vision}, pages 2556--2563, 2011.
\newblock \doi{10.1109/ICCV.2011.6126543}.

\bibitem[Yu et~al.(2018)Yu, Wang, Zhao, and Ji]{yu2018fine}
Huanyu Yu, Shuo Wang, Zhou Zhao, and Wen Ji.
\newblock Fine-grained video captioning for sports narrative.
\newblock In \emph{Proceedings of the IEEE Conference on Computer Vision and
  Pattern Recognition}, pages 6006--6015, 2018.
\newblock \doi{10.1109/CVPR.2018.00628}.

\bibitem[Qi et~al.(2019)Qi, Wang, Li, and Luo]{qi2019sports}
Meng Qi, Yihan Wang, Anyi Li, and Jiebo Luo.
\newblock Sports video captioning via attentive motion representation and group
  relationship modeling.
\newblock \emph{IEEE Transactions on Circuits and Systems for Video
  Technology}, 30\penalty0 (8):\penalty0 2617--2633, 2019.
\newblock \doi{10.1109/TCSVT.2019.2924187}.

\bibitem[Xu et~al.(2022)Xu, Rao, Yu, Chen, Zhou, and Lu]{xu2022finediving}
Jinglin Xu, Yongming Rao, Xumin Yu, Guangyi Chen, Jie Zhou, and Jiwen Lu.
\newblock Finediving: A fine-grained dataset for procedure-aware action quality
  assessment.
\newblock In \emph{Proceedings of the IEEE/CVF Conference on Computer Vision
  and Pattern Recognition}, pages 2949--2958, 2022.
\newblock \doi{10.1109/CVPR52688.2022.00295}.

\bibitem[Zhu et~al.(2018)Zhu, Lan, Newsam, and Hauptmann]{zhu2018hidden}
Yi~Zhu, Zhenzhong Lan, Shawn Newsam, and Alexander Hauptmann.
\newblock Hidden two-stream convolutional networks for action recognition.
\newblock In \emph{Asian Conference on Computer Vision}, pages 363--378.
  Springer, 2018.
\newblock \doi{10.1007/978-3-030-20873-8_23}.

\bibitem[Zhang et~al.(2016)Zhang, Wang, Wang, Qiao, and Wang]{zhang2016real}
Bowen Zhang, Limin Wang, Zhe Wang, Yu~Qiao, and Hanli Wang.
\newblock Real-time action recognition with enhanced motion vector cnns.
\newblock In \emph{Proceedings of the IEEE Conference on Computer Vision and
  Pattern Recognition}, pages 2718--2726, 2016.
\newblock \doi{10.1109/CVPR.2016.297}.

\bibitem[Vala and Baxi(2021)]{vala2021key}
Hardik~J Vala and Astha Baxi.
\newblock Key frame extraction of surveillance video based on frequency domain
  analysis.
\newblock \emph{Intelligent Automation \& Soft Computing}, 29\penalty0
  (1):\penalty0 145--160, 2021.
\newblock \doi{10.32604/iasc.2021.017772}.

\bibitem[Tint and Soe(2020)]{tint2020key}
Myat~Su Tint and Thae~Thae Soe.
\newblock Key frame extraction for video summarization using dwt wavelet.
\newblock In \emph{2020 IEEE Conference on Computer Applications}, pages 1--6.
  IEEE, 2020.
\newblock \doi{10.1109/ICCA49400.2020.9022814}.

\bibitem[Kodathala et~al.(2025)Kodathala, Vutukoori, and
  Vunnam]{kodathala2025sv33bsportsvideounderstanding}
Sai~Varun Kodathala, Yashwanth~Reddy Vutukoori, and Rakesh Vunnam.
\newblock Sv3.3b: A sports video understanding model for action recognition,
  2025.
\newblock URL \url{https://arxiv.org/abs/2507.17844}.

\bibitem[Kapre et~al.(2023)Kapre, Rajurkar, and Guru]{kapre2023improved}
Bhavana~S Kapre, Anand~M Rajurkar, and DS~Guru.
\newblock An improved video keyframe detection technique leads to video
  authentication.
\newblock \emph{International Journal of Engineering Trends and Technology},
  71\penalty0 (4):\penalty0 171--189, 2023.
\newblock \doi{10.14445/22315381/IJETT-V71I4P218}.

\bibitem[Singh and Raj(2025)]{singh2025key}
Ankit~Kumar Singh and Bharat Raj.
\newblock Key frame extraction algorithm for surveillance videos using an
  improved genetic algorithm.
\newblock \emph{Scientific Reports}, 15\penalty0 (1):\penalty0 1--15, 2025.
\newblock \doi{10.1038/s41598-024-85421-2}.

\bibitem[Zhuang et~al.(1998)Zhuang, Rui, Huang, and
  Mehrotra]{zhuang1998adaptive}
Yueting Zhuang, Yong Rui, Thomas~S Huang, and Sharad Mehrotra.
\newblock Adaptive key frame extraction using unsupervised clustering.
\newblock In \emph{Proceedings of 1998 International Conference on Image
  Processing}, volume~1, pages 866--870. IEEE, 1998.
\newblock \doi{10.1109/ICIP.1998.723652}.

\bibitem[Karpathy et~al.(2014)Karpathy, Toderici, Shetty, Leung, Sukthankar,
  and Fei-Fei]{karpathy2014large}
Andrej Karpathy, George Toderici, Sanketh Shetty, Thomas Leung, Rahul
  Sukthankar, and Li~Fei-Fei.
\newblock Large-scale video classification with convolutional neural networks.
\newblock In \emph{Proceedings of the IEEE Conference on Computer Vision and
  Pattern Recognition}, pages 1725--1732, 2014.
\newblock \doi{10.1109/CVPR.2014.223}.

\bibitem[Ng et~al.(2015)Ng, Hausknecht, Vijayanarasimhan, Vinyals, Monga, and
  Toderici]{ng2015beyond}
Joe Yue-Hei Ng, Matthew Hausknecht, Sudheendra Vijayanarasimhan, Oriol Vinyals,
  Rajat Monga, and George Toderici.
\newblock Beyond short snippets: Deep networks for video classification.
\newblock In \emph{Proceedings of the IEEE Conference on Computer Vision and
  Pattern Recognition}, pages 4694--4702, 2015.
\newblock \doi{10.1109/CVPR.2015.7299101}.

\bibitem[Hara et~al.(2018)Hara, Kataoka, and Satoh]{hara2018can}
Kensho Hara, Hirokatsu Kataoka, and Yutaka Satoh.
\newblock Can spatiotemporal 3d cnns retrace the history of 2d cnns and
  imagenet?
\newblock In \emph{Proceedings of the IEEE Conference on Computer Vision and
  Pattern Recognition}, pages 6546--6555, 2018.
\newblock \doi{10.1109/CVPR.2018.00685}.

\bibitem[Feichtenhofer et~al.(2019)Feichtenhofer, Fan, Malik, and
  He]{feichtenhofer2019slowfast}
Christoph Feichtenhofer, Haoqi Fan, Jitendra Malik, and Kaiming He.
\newblock Slowfast networks for video recognition.
\newblock In \emph{Proceedings of the IEEE/CVF International Conference on
  Computer Vision}, pages 6202--6211, 2019.
\newblock \doi{10.1109/ICCV.2019.00630}.

\bibitem[van~der Maaten and Hinton(2008{\natexlab{a}})]{maaten2008visualizing}
Laurens van~der Maaten and Geoffrey Hinton.
\newblock Visualizing data using t-sne.
\newblock \emph{Journal of Machine Learning Research}, 9:\penalty0 2579--2605,
  2008{\natexlab{a}}.

\bibitem[Alemi et~al.(2017)Alemi, Fischer, Dillon, and Murphy]{alemi2017deep}
Alexander~A Alemi, Ian Fischer, Joshua~V Dillon, and Kevin Murphy.
\newblock Deep variational information bottleneck.
\newblock In \emph{International Conference on Learning Representations}, 2017.

\bibitem[Tran et~al.(2018)Tran, Wang, Torresani, Ray, LeCun, and
  Paluri]{tran2018closer}
Du~Tran, Heng Wang, Lorenzo Torresani, Jamie Ray, Yann LeCun, and Manohar
  Paluri.
\newblock A closer look at spatiotemporal convolutions for action recognition.
\newblock In \emph{Proceedings of the IEEE Conference on Computer Vision and
  Pattern Recognition}, pages 6450--6459, 2018.
\newblock \doi{10.1109/CVPR.2018.00675}.

\bibitem[Wang et~al.(2018)Wang, Girshick, Gupta, and He]{wang2018non}
Xiaolong Wang, Ross Girshick, Abhinav Gupta, and Kaiming He.
\newblock Non-local neural networks.
\newblock In \emph{Proceedings of the IEEE Conference on Computer Vision and
  Pattern Recognition}, pages 7794--7803, 2018.
\newblock \doi{10.1109/CVPR.2018.00813}.

\bibitem[Laptev(2005)]{laptev2005space}
Ivan Laptev.
\newblock On space-time interest points.
\newblock \emph{International Journal of Computer Vision}, 64\penalty0
  (2-3):\penalty0 107--123, 2005.
\newblock \doi{10.1007/s11263-005-1838-7}.

\bibitem[Chen et~al.(2020)Chen, Kornblith, Norouzi, and Hinton]{chen2020simple}
Ting Chen, Simon Kornblith, Mohammad Norouzi, and Geoffrey Hinton.
\newblock A simple framework for contrastive learning of visual
  representations.
\newblock In \emph{International Conference on Machine Learning}, pages
  1597--1607. PMLR, 2020.

\bibitem[Simonyan and Zisserman(2015)]{simonyan2014very}
Karen Simonyan and Andrew Zisserman.
\newblock Very deep convolutional networks for large-scale image recognition.
\newblock In \emph{International Conference on Learning Representations}, 2015.

\bibitem[Sim{\'e}oni et~al.(2025)Sim{\'e}oni, Vo, Seitzer, Baldassarre, Oquab,
  Jose, Khalidov, Szafraniec, Yi, Ramamonjisoa, Massa, Haziza, Wehrstedt, Wang,
  Darcet, Moutakanni, Sentana, Roberts, Vedaldi, Tolan, Brandt, Couprie,
  Mairal, J{\'e}gou, Labatut, and Bojanowski]{simeoni2025dinov3}
Oriane Sim{\'e}oni, Huy~V. Vo, Maximilian Seitzer, Federico Baldassarre, Maxime
  Oquab, Cijo Jose, Vasil Khalidov, Marc Szafraniec, Seungeun Yi, Micha{\"e}l
  Ramamonjisoa, Francisco Massa, Daniel Haziza, Luca Wehrstedt, Jianyuan Wang,
  Timoth{\'e}e Darcet, Th{\'e}o Moutakanni, Leonel Sentana, Claire Roberts,
  Andrea Vedaldi, Jamie Tolan, John Brandt, Camille Couprie, Julien Mairal,
  Herv{\'e} J{\'e}gou, Patrick Labatut, and Piotr Bojanowski.
\newblock {DINOv3}, 2025.
\newblock URL \url{https://arxiv.org/abs/2508.10104}.

\bibitem[van~der Maaten and
  Hinton(2008{\natexlab{b}})]{vandermaaten2008visualizing}
Laurens van~der Maaten and Geoffrey Hinton.
\newblock Visualizing data using t-sne.
\newblock \emph{Journal of Machine Learning Research}, 9:\penalty0 2579--2605,
  2008{\natexlab{b}}.

\bibitem[McInnes et~al.(2018)McInnes, Healy, and Melville]{mcinnes2018umap}
Leland McInnes, John Healy, and James Melville.
\newblock Umap: Uniform manifold approximation and projection for dimension
  reduction, 2018.

\end{thebibliography}






\end{document}